\documentclass{article}

\PassOptionsToPackage{numbers, compress}{natbib}
\usepackage[preprint]{style_templete}

\usepackage[utf8]{inputenc} % allow utf-8 input
\usepackage[T1]{fontenc}    % use 8-bit T1 fonts
\usepackage{hyperref}       % hyperlinks
\usepackage{url}            % simple URL typesetting
\usepackage{booktabs}       % professional-quality tables
\usepackage{nicefrac}       % compact symbols for 1/2, etc.
\usepackage{microtype}      % microtypography
\usepackage{enumitem}

\usepackage{algorithm}% http://ctan.org/pkg/algorithms
\usepackage{xcolor}
\usepackage{algorithmic}

\usepackage{graphicx,subfigure}
\usepackage{amsmath,amsfonts}
\usepackage{multirow,soul}

\usepackage{placeins}
\usepackage{tabularx,threeparttable}
\usepackage{algorithm}
\usepackage{color}
\usepackage{epstopdf}
\usepackage{amsmath,amssymb}
\usepackage{setspace}

\usepackage{placeins}
\usepackage{tabularx,threeparttable}
\usepackage{algorithm}
\usepackage{color}
\usepackage{epstopdf}
\usepackage{amsmath,amssymb}
\usepackage{setspace}
\usepackage{HJnotations}
\usepackage{mathrsfs}

\definecolor{purple}{RGB}{250,000,180}

\def\blue{\color{blue}}

\definecolor{maroon}{RGB}{122,000,25}

\title{LATTE: Label-efficient Incident Phenotyping from Longitudinal Electronic Health Records}

\author{  \\
Jun Wen\textsuperscript{1,2}, Jue Hou\textsuperscript{3}, Clara-Lea Bonzel\textsuperscript{1,2},
Yihan Zhao\textsuperscript{4},
Victor M. Castro \textsuperscript{5},  \\
Vivian S. Gainer \textsuperscript{5},  Dana Weisenfeld\textsuperscript{6}, Tianrun Cai\textsuperscript{2,5}, Yuk-Lam Ho\textsuperscript{2}, \\
Vidul A. Panickan\textsuperscript{1,2}, Lauren Costa\textsuperscript{2},  Chuan Hong\textsuperscript{7}, \\
J. Michael Gaziano\textsuperscript{1,2,6}, Katherine P. Liao\textsuperscript{1,2,6},  Junwei Lu\textsuperscript{2,8},  \\
Kelly Cho\textsuperscript{1,2,6}, Tianxi Cai\textsuperscript{1,2,8}\thanks{Corresponding author (tcai@hsph.harvard.edu). }  \\
\textsuperscript{1} Harvard Medical School, Boston, MA, USA \\
\textsuperscript{2} VA Boston Healthcare System, Boston, MA, USA \\
\textsuperscript{3} University of Minnesota, Minneapolis, MN, USA \\
\textsuperscript{4}  Harvard University, Cambridge, MA, USA \\
\textsuperscript{5} Mass General Brigham, Boston, MA, USA \\
\textsuperscript{6} Brigham and Women’s Hospital, Boston, MA, USA \\
\textsuperscript{7} Duke University, Durham, NC, USA \\
\textsuperscript{8} Harvard T.H. Chan School of Public Health, Boston, MA, USA }

\begin{document}

\maketitle

\textbf{Highlights}:
\begin{itemize}
\item A framework, called LATTE, is proposed to allow label-efficient incident phenotyping from longitudinal electronic health records.
\item Incident-indicative concepts and visits are learned efficiently by exploiting semantic embeddings learned from large-scale EHR data as prior knowledge.
\item Longitudinal silver-standard labels built upon predictive surrogates alleviate dependency on gold-standard labels substantially.
\item Significant performance boost over existing methods are achieved consistently in identifying incident type-2 diabetes, heart failure, and multiple sclerosis while allowing high prediction interpretability and cross-site portability.
\item Accurate incident phenotyping from LATTE facilitates efficient downstream analysis on the time-to-event outcome, which is exemplified by discovering risk factors of heart failure among patients with rheumatoid arthritis.
\end{itemize}

% \begin{figure*}[!hbtp]
% \begin{center}
% \includegraphics[width= \textwidth]{imgs/graphic abstract.pdf}
%    \caption{Graphic abstract.}
%    \label{figure:abstract}
% \end{center}
% \end{figure*}

\newpage

\begin{abstract}
 \textbf{Objective:}
 Electronic health record (EHR) data are increasingly used to support real-world evidence (RWE) studies. Yet its ability to generate reliable RWE is limited by the lack of readily available precise information on the timing of clinical events such as the onset time of heart failure. Rule-based algorithms tend to have limited accuracy while supervised machine learning algorithms typically require a large number of gold standard labels, which is resource intensive to obtain. We propose a LAbel-efficienT incidenT phEnotyping (LATTE) algorithm to accurately annotate the timing of clinical events from longitudinal EHR data.

 \textbf{Methods:}
By leveraging the pre-trained semantic embedding vectors from large-scale EHR data as prior knowledge, LATTE selects predictive EHR features in a concept re-weighting module by mining  their relationship to the target event and compresses their information into longitudinal visit embeddings through a visit attention learning network.
LATTE employs a recurrent neural network to capture the sequential dependency between the target event and visit embeddings before/after it. To improve label efficiency, LATTE constructs highly informative longitudinal silver-standard labels from large-scale unlabeled patients to perform unsupervised pre-training and semi-supervised joint training. Finally, LATTE enhances cross-site portability via contrastive representation learning.

\textbf{Results:}
LATTE is evaluated on three analyses: the onset of type-2 diabetes, heart failure, and the onset and relapses of multiple sclerosis.
We use various evaluation metrics present in the literature including the $ABC_{gain}$, the proportion of reduction in the area between the observed event indicator and the predicted cumulative incidences in reference to the prediction per incident prevalence. LATTE consistently achieves substantial improvement over benchmark methods such as SAMGEP and RETAIN in all settings. For example, it outperforms SAMGEP by 11.0\% on average in terms of $ABC_{gain}$ when only 100 gold-standard labels are available. Remarkably, when trained only using the proposed longitudinal silver labels, LATTE achieves a performance comparable to deep learning models trained using hundreds of gold-standard labels. Furthermore, LATTE provides high interpretation capability by indicating which concepts at which EHR visits contribute to the incident prediction. Finally, we show that the precise timing of incidents provided by LATTE facilitates the discovery of risk factors for heart failure among patients with rheumatoid arthritis.

%LATTE consists of three steps: (i) learn concept weight based the semantic relationship to the target phenotype, (ii) learn discriminative attentions to longitudinal visits, and (iii) employs recurrent neural networks to model visit sequential dependency and localize incidents. To achieve high-efficiency in utilization of gold-standard labels, LATTE constructs longitudinal silver labels upon predictive surrogates for feature selection, pre-training and co-training. LATTE strengthens cross-site generalizability via constrastive representation learning on manipulated gold-standard annotations.

 %LATTE is built on EHR concept embedding and designed with high efficiency in utilization of gold-standard labels. It mainly composed of five parts. (i) Based on concept embedding, it discriminatively distinguishes the importance of input concepts via a concept re-weighting module. (ii) It employs a self-attention technique to pay attention to important visits and prevent influences from noise visits. (iii) It utilizes recurrent neural networks to model the sequential dependency among visits and identify incident timings. (iv) For high-efficiency in utilizing gold-standard labels, LATTE performs unsupervised representation learning and co-training by constructing longitudinal silver labels. (v) LATTE strengthens cross-site generalization by performing contrastive representation learning on additionally augmented patient data.

\end{abstract}

% \hj{First paragraph needs to be expanded. 1) great promise of EHR, list some FDA papers
% supporting EHR research; 2) binary phenotyping has facilitated many studies (summarize the applications, e. g. genome-phenom association); 3) bring out the event time: timing has richer information than binary status, accounting for censoring (heterogeneity in follow-up).
% Start a new paragraph to lay out the background of incidence phenotyping: where is the information and where is the noise?}

\section{Introduction}
In recent years, electronic health record (EHR) data collected during the routine delivery of care has opened opportunities for discovery and translational research \cite{kohane2012translational,miotto2016deep}. For example, EHR-derived cohorts have led to large-scale clinical studies and phenome-wide association studies  \cite{murphy2006integration,liao2010electronic,ananthakrishnan2013improving,roden2008development}. Due to their large size and broad patient population, EHR cohorts are increasingly used to support real-world evidence (RWE) on the efficacy and safety of therapeutic drugs or intervention procedures \cite{gamerman2019pragmatic,hernandez2019real,hou2022temporal,hou2021comparison}.
 However, the capacity of EHR data for supporting RWE studies is currently limited due to the lack of direct observations on the precise timing of clinical events such as the onset of heart failure.
The timing information plays an important  role in RWE studies, including in determining eligibility at baseline or defining time-to-event outcomes \cite{huang2021association}.
 Readily available EHR features such as the timing of relevant international classification of disease (ICD) codes are often inaccurate due to either miscoding or ICD codes being assigned to visits that rule out a disease. Additionally, event-time derived surrogates tend to have systematic biases \cite{uno2018determining,hassett2017detecting}. On the other hand, it is time and resource-prohibitive to extract event information via manual chart review. For binary phenotype traits such as the presence or absence of a condition, a wide range of supervised, unsupervised, and label-efficient semi-supervised machine learning-based phenotyping algorithms have been successfully developed and validated across many disease phenotypes \cite{liao2019high,yu2018enabling,ahuja2020surelda,kirby2016phekb,newton2013validation,ocac216}.
On the contrary, few methods currently exist to accurately and efficiently derive computational event time phenotypes based on longitudinal EHR data.

 Existing approaches to deriving computational event time phenotypes can generally be categorized as rule-based \cite{chubak2012administrative,uno2018determining} and machine-learning-based \cite{choi2016retain,ahuja2021samgep}. For example, Chubak et al. developed rules to predict breast cancer recurrence based on the earliest observation of expert-specified codes \cite{chubak2012administrative}. Uno et al. proposed to alleviate the systematic temporal biases between code timings and phenotype onset by using points of maximal increase in lieu of peak values \cite{uno2018determining}.  Even though rule-based methods can achieve notable performance for some  phenotypes, they are limited by the reliance on expert knowledge to curate a small set of predictive surrogate concepts, which prevents their application to diseases without such predictive concepts or  the scaling-up to data with hundreds of unspecific features. Rule-based methods are tailored to specific applications such as cancer recurrence per domain knowledge and are hardly generalizable to other applications.

A more generalizable alternative approach is to employ machine learning in order to derive computational incident phenotyping algorithms using temporal patterns of EHR data. For example, random forests were investigated for phenotyping opioid overdose events \cite{badger2019machine}. Due to the stronger capacity in capturing temporal dependency, sequential models are prevalently introduced.
In this line, deep neural networks have been studied for event-time phenotyping. For example, a graph-based framework is proposed for temporal phenotyping by Liu et al. \cite{liu2015temporal}, and recurrent neural
networks are employed to mimic physician attentions for interpretable phenotyping \cite{choi2016retain} and to perform outcome-oriented temporal phenotyping \cite{lee2020outcome}. However, to combat overfitting, such deep learning-based algorithms depend on a large number of labels that are expensive to obtain and not widely available. Recently, a semi-supervised algorithm, SAMGEP \cite{ahuja2021samgep}, is developed to model disease progression as a Gaussian Process emission through a Hidden Markov model (HMM). Despite the high label efficiency, SAMGEP imposes relatively simple linear effects, which limits its ability to capture complex sequential dependency patterns.

In this paper, we propose a semi-supervised LAbel-efficienT incidenT phEnotyping (LATTE) algorithm to derive the timing of clinical incidents from longitudinal EHR data. LATTE attains high accuracy with a small label size by effectively leveraging longitudinal silver-standard labels and the prior knowledge from semantic embeddings of EHR concepts to perform unsupervised pre-training and semi-supervised model co-training. Another key advantage of LATTE compared to existing literature lies in its cross-site portability, which is enabled via contrastive representation learning. Efficient and accurate annotations of clinical event times through LATTE strengthen the potential of EHR data for generating real-world evidence.

\section{Results}
\label{sec_experiments}
We first evaluate LATTE on three representative phenotypes to demonstrate its advantages over existing methods in label efficiency and cross-site portability for incident phenotyping. Further, based on phenotype incident predictions on longitudinal EHR data, we identify risk factors of heart failure among patients with rheumatoid arthritis (RA).

\subsection{Performance of Incident Phenotyping}
\subsubsection{Data and Settings}
\label{sec_experiment_setting}
\noindent \textbf{Data Sources:} We first evaluate the performance of LATTE in identifying three temporal events, the onset of Type-2 diabetes (T2D, PheCode 250.2) and heart failure (HF, PheCode 428), and the relapses over time for those with Multiple sclerosis (MS, PheCode 335), using EHR data from Mass General Brigham (MGB). Both T2D and HF are chronic diseases for which the first onset probability over time is of primary interest.  MS relapse is a relapsing and remitting phenotype for which we aim to predict all recurrent relapses over time. For both T2D and HF, the corresponding diagnostic codes are predictive of the ever/never status, but the dates of the first diagnostic codes often deviate from the true incident times with systematic preceding and lagging biases. For MS relapse, no predictive diagnostic code exists, so complex visit dependency modeling is required to precisely identify incident visits.
For HF, we further evaluate the transportability of the algorithm to a Biobank cohort at MGB and to the million veteran project (MVP) cohort at VA.

For T2D diabetes, we assemble EHR data for 10,315 patients who have at least one T2D ICD code from the Mass General Brigham (MGB) healthcare system. The T2D status and onset dates for 172 randomly selected patients from this cohort were annotated via chart review. Among the 172 patients,  52.3\%  develop T2D during follow-up and 11.0\%  develop T2D before the first EHR visit. 10-fold cross-validation is used for performance evaluation. For MS relapse, we assemble EHR data for 4,706 patients at MGB, out of which 1,435 patients are participants of the Comprehensive Longitudinal Investigation of Multiple Sclerosis at BWH (CLIMB) research registry with relapse status annotated over time.
Within the CLIMB cohort, 57.2\% of patients have at least one relapse event, with a mean of 2.60 relapses per patient.

For HF, we train the algorithm using EHR data from the MGB RA cohort in which each patient has at least 1 ICD code for rheumatoid arthritis.
HF status and onset time are annotated for a random subset of 234 patients, among which 60.7\% are determined to have developed HF during follow-up. Beyond 10-fold cross-validation, We further evaluate the portability of the HF incident phenotyping algorithm trained in the MGB RA cohort to the MGB Biobank cohort and the million veteran project (MVP) cohort at the VA. The MGB Biobank and VA-MVP cohorts consist of 13597 and 122035 patients with at least 1 ICD code of HF, out of which 94 and 208 are randomly annotated and used for transportability validation.

We bin the longitudinal EHR data into consecutive, non-overlapping 3-month time windows \cite{ahuja2021samgep}. For T2D, we use the codified features selected by KESER \cite{hong2021clinical}. For HF, the medical notes are found to be important to determine the disease onset, and thus NLP CUIs relevant to HF are extracted from medical notes as candidate features. We select the HF-related CUIs via the CUIs search tool \footnote{http://app.parse-health.org/CUISearch/}. For MS relapse, 155 EHR features are manually selected by a domain expert as in a previous study \cite{ahuja2021samgep}.

\noindent \textbf{Compared methods:} We consider three benchmark methods: (i) long short term memory RNN (LSTM), (ii) RETAIN which is trained with longitudinal raw EHR features and without effective utilization of pre-trained concept embedding vectors, and (iii) SAMGEP which aggregates patient visit embedding as a pre-processing step, without learning to distinguish the importance of concepts/visits, and mines only linear and feed-forward dependencies between visits, lacking the capacity to model the complex and long-range temporal relationship between incidents and patient visits.
For both methods, the training of incidence phenotyping models depends exclusively on the labeled data with gold-standard outcomes, overlooking the information from  the predictive surrogates of the vast unlabeled data.
As a baseline, we also include predictions based only on the closest PheCodes (T2D: 250.2; HF: 428; MS relapse: 355;) and PheNorm with temporally cumulative counts \cite{yu2018enabling}, denoted as PheNorm(acc), which is shown to perform effectively on ever/never phenotyping.

\noindent \textbf{Evaluation measures:} To evaluate the methods’ performance, we sample various sizes of patients from the gold-standard labeled set to train the model and evaluate each trained model on the rest of the labels. To quantify the accuracy of the methods' predictions, we compute (i) AUC (Area Under the Receiver Operating Characteristic Curve), and (ii) F1 score, with a cutoff value that achieves 95\% specificity. The two values measure the incident identification error by treating each visit independently.
We also report the methods' longitudinal phenotype predictions, namely the area between the label curve $Y_i^t$ and predicted cumulative probability $\hat{Y}_i^t=1-\prod_{k=0}^{k \leq t}(1-p_i^k)$, denoted by $ABC_{cdf}$, where $p_i^k$ denotes patient $i$'s prediction at time $k$. $ABC_{cdf}$ effectively evaluates the mean absolute difference between true and predicted incident times but would scale up when patients with longer EHR observations. (iii) Therefore, we compute the normalized version, $ABC_{gain}$, which is the methods’ percent decrease over a $null$ model that sets the probability at each visit to the prevalence of the phenotypes \cite{ahuja2021samgep}, namely $ABC_{gain}=(ABC_{cdf,null}-ABC_{cdf,method})/ABC_{cdf,null}$.

\subsubsection{Result Analysis}

\begin{figure*}[tb]
  \centering
  \subfigure[T2D]{\includegraphics[scale=0.42]{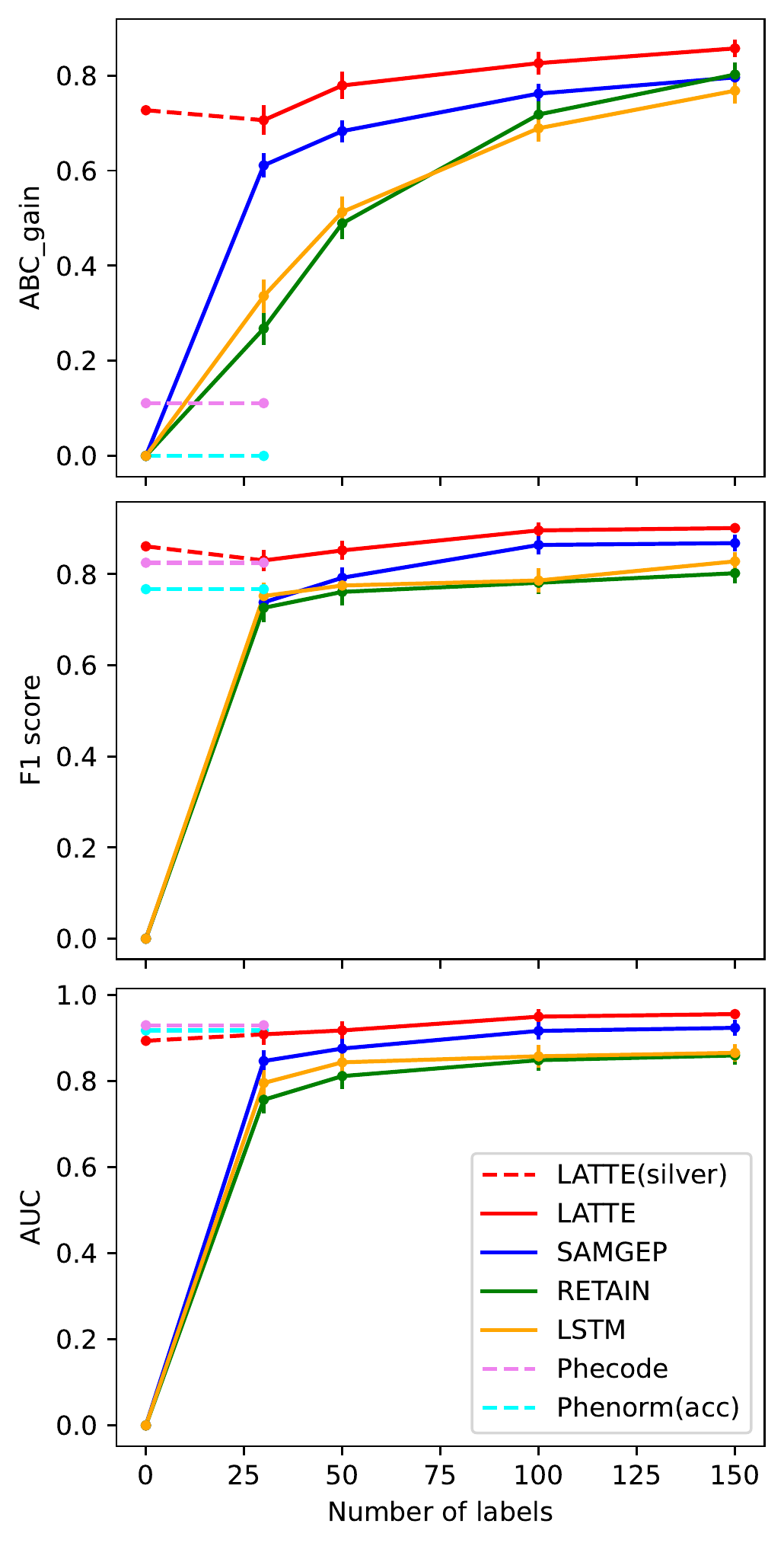}}
  \subfigure[HF]{\includegraphics[scale=0.42]{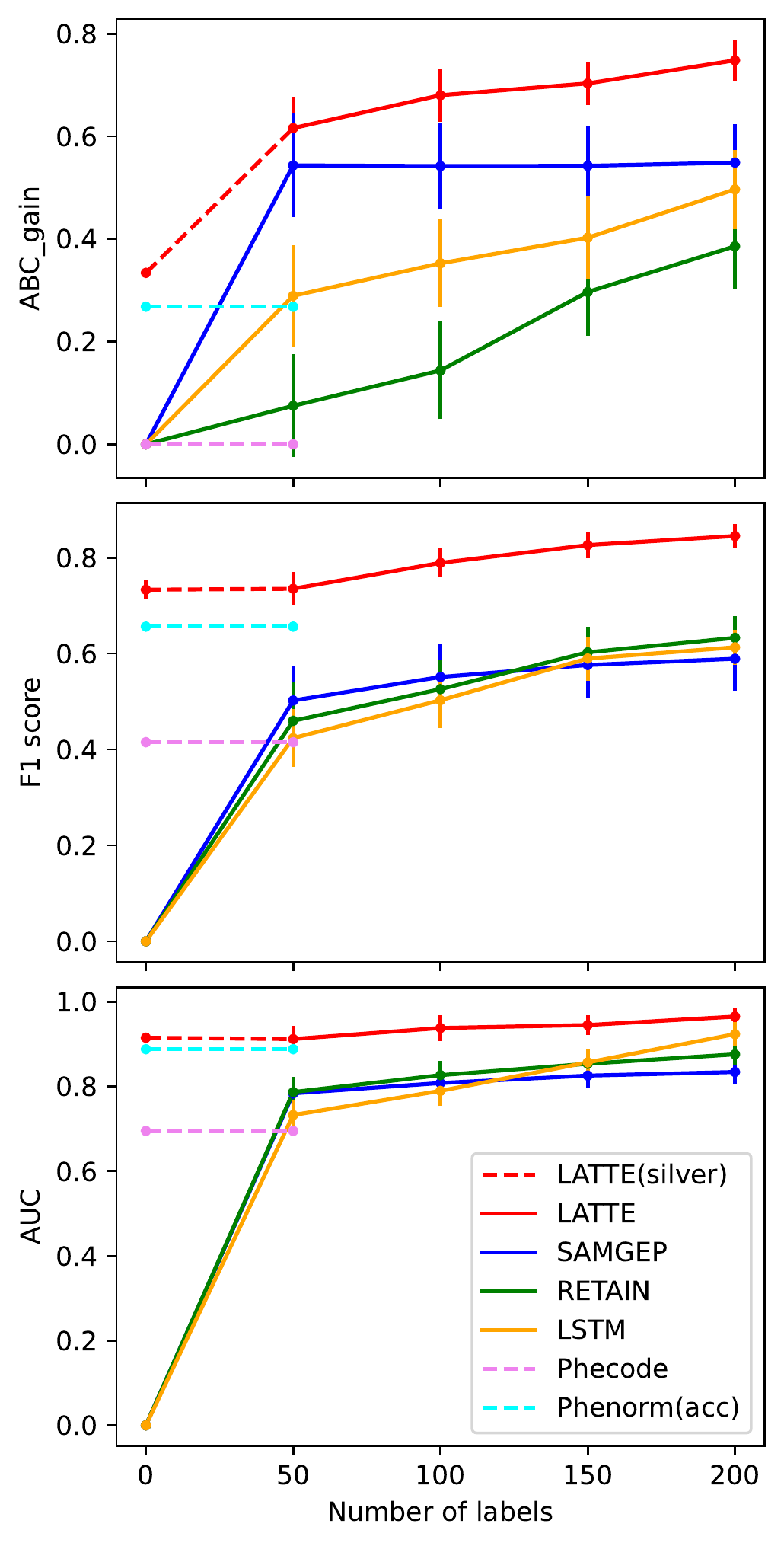}}
  \subfigure[MS]{\includegraphics[scale=0.42]{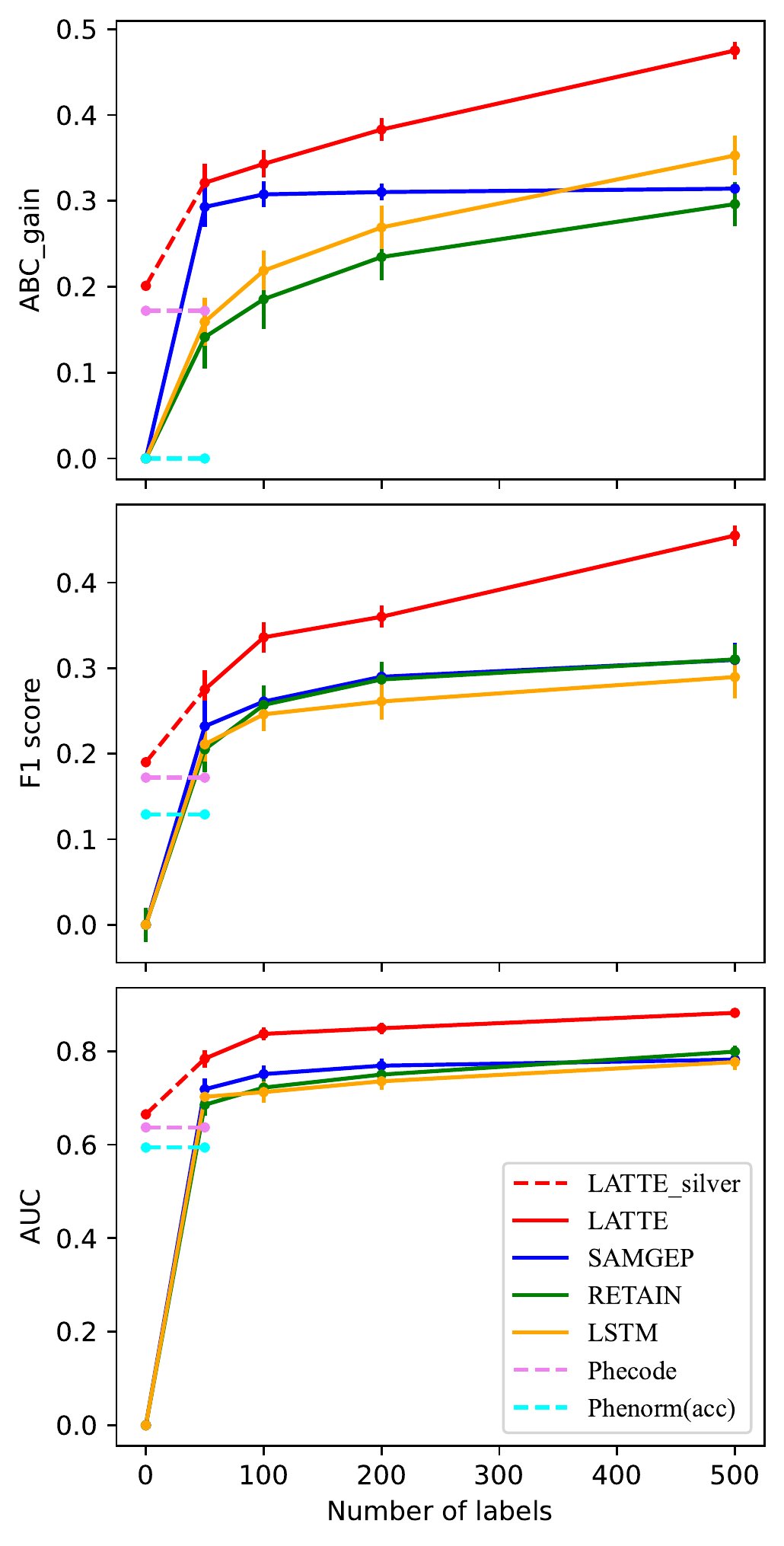}}
  \caption{Results of Incident phenotyping on type-2 diabetes (T2D), heart failure (HF), and multiple sclerosis (MS) relapses. The dashed lines denote unsupervised methods that do not use any gold-standard labels.}
  \label{fig:results_incidents}
\end{figure*}

The phenotyping results are shown in Figure \ref{fig:results_incidents} with 95\%-interval error bars. For better visualization, we truncate $ABC_{gain}$ values to be above 0. Different sizes of gold-standard label sets are evaluated and the ``0'' labels, visualized in dashed lines, denote the unsupervised scenario where no gold-standard label is used.

\noindent \textbf{Type-2 diabetes:} The results of classifying type-2 diabetes' first onset time in the MGB cohort are shown in Figure \ref{fig:results_incidents} (a). LATTE shows significant advantages over the compared methods in terms of $ABC_{gain}$, F1 score, and AUC. The unsupervised approaches, PheCode only, PheNorm(ACC), and LATTE(silver) achieve comparable performance in terms of AUC and F1 score compared to the supervised models, LSTM, RETAIN, SAMGEP, and LATTE. This indicates that the main PheCode itself is able to well distinguish the case visits from control visits. In addition to that, the PheCode and PheNorm(acc) are significantly outperformed in $ABC_{gain}$ by the supervised models, which indicates that PheCode only fails to precisely localize the incidents. On the other hand, the performance of LATTE(silver), which is without gold-standard labels, is comparable to LSTM, RETAIN, and SAMGEP, trained by 100 gold-standard labels, which shows the promise of unsupervised representation learning upon the predictive surrogate concepts. Of note is, although LATTE(silver) exploits the main PheCode as a strong surrogate, which is the same as in the PheNorm, it optimizes the sequential representation learned by GRUs to output non-decreasing incident risks across time.
As a result, the predictions obtained using LATTE(silver) are guided to be more sensitive to the onsets of type-2 diabetes. When only small sets of labels are available, for example, 30 or 50, the deep learning models LSTM and RETAIN suffer severe over-fitting and are outperformed by SAMGEP in $ABC_{gain}$ while the performance gaps become smaller as more gold-standard labels are available.

\noindent \textbf{Heart failure:} The results of predicting heart failure's first onset on the MGB RA cohort are provided in Figure \ref{fig:results_incidents} (b). LATTE(silver) consistently outperforms the LSTM and RETAIN when the label set size is small. SAMGEP achieves comparable performance to  RETAIN and LSTM in terms of AUC and F1 scores but shows significant improvements in $ABC_{gain}$ which becomes minor as more labels are available. With increasing label set size, the improvements from SAMGEP are limited but are significant from LATTE, LSTM, and RETAIN, and the performance gap between LATTE and SAMGEP trows, which justifies the stronger learning capability of LATTE than SAMGEP.

\noindent \textbf{Multiple sclerosis relapse:} The results of localizing multiple sclerosis's relapses are provided in Figure \ref{fig:results_incidents} (c).
Identifying MS relapse incidents is more challenging than the first onset of T2D and HF as MS relapse can not be well captured by a simple code or NLP concept and thus does not have a highly predictive surrogate concept.
We therefore further increase the size label set size to 500 for comprehensive evaluation. LATTE(silver) is only comparable to the PheCodes, and both methods are significantly outperformed by the supervised models with 100 labels or more, which indicates that the PheCodes are non-predictive of the incidents. When available labels are less than 200, SAMGEP outperforms the LSTM and RETAIN while when label size grows to 500, SAMGEP is narrowly outperformed by the LSTM in $ABC_{gain}$ and the performance advantages of LATTE over LSTM, RETAIN, and SAMGEP become more distinct, especially in $ABC_{gain}$.

The results of transportability to MGB Biobank and VA-MVP for HF incident phenotyping algorithm trained at MGB RA cohort are provided in Table \ref{Table:HF_cross_site}. Both LSTM and RETAIN methods outperform the PheCode:428 in terms of $ABC_{gain}$, although the PheCode-only is distinctive of case/control visits and has high AUC and F1 scores on the Biobank-HF. LATTE suffers an average performance drop in  $ABC_{gain}$ by 11.9\% on the Biobank-HF and VA-MVP, which is significantly better than the $66.1\%$ of LSTM and $25.6\%$ of RETAIN.

\begin{table}[]
  \centering
  \caption{Cross-site validation of incident phenotyping on heart failure from MGB RA cohort (10-fold cross-validation performance reported) to Biobank-HF and VA MVP.}
\begin{tabular}{c|rrrr}
\toprule[1pt]
    Settings & Method  & AUC & F1 & $ABC_{gain}$ \\
\hline
  \multirow{1}*{\textcolor{red}{MGB-RA} $\rightarrow$ \blue{Biobank-HF}}    & PheCode   &  0.915 &  0.776  &-4.83    \\
                                                  & LSTM           &\textcolor{red}{0.928} $\rightarrow$ \blue{0.689}  & \textcolor{red}{0.669} $\rightarrow$ \blue{0.588}   & \textcolor{red}{0.635} $\rightarrow$ \blue{0.105}    \\
                                                  & RETAIN         &\textcolor{red}{0.889} $\rightarrow$ \blue{0.743}  & \textcolor{red}{0.695} $\rightarrow$ \blue{0.640}   &\textcolor{red}{0.445} $\rightarrow$ \blue{0.243}    \\
                                                  & LATTE          &\textcolor{red}{0.969} $\rightarrow$ \blue{0.879}  & \textcolor{red}{0.850} $\rightarrow$ \blue{0.790}   & \textcolor{red}{0.752} $\rightarrow$ \blue{0.675}  \\

   \hline
   \multirow{1}*{\textcolor{red}{MGB-RA} $\rightarrow$ \blue{VA-MVP}}      & PheCode   &  0.663&  0.478    &0.289     \\
                                             & LSTM           &\textcolor{red}{0.928} $\rightarrow$ \blue{0.745}  & \textcolor{red}{0.669} $\rightarrow$ \blue{0.638}     & \textcolor{red}{0.635} $\rightarrow$ \blue{0.325}     \\
                                             & RETAIN         &\textcolor{red}{0.889} $\rightarrow$ \blue{0.785}  & \textcolor{red}{0.695} $\rightarrow$ \blue{0.667}     & \textcolor{red}{0.445} $\rightarrow$ \blue{0.419}     \\
                                             & LATTE          &\textcolor{red}{0.969} $\rightarrow$ \blue{0.892}  & \textcolor{red}{0.850} $\rightarrow$ \blue{0.765}     & \textcolor{red}{0.752} $\rightarrow$ \blue{0.650}    \\
\toprule[1pt]
\end{tabular}
\label{Table:HF_cross_site}
\end{table}

\begin{figure*}[tb]
  \centering
  \subfigure[Raw (T2D)]{\includegraphics[scale=0.32]{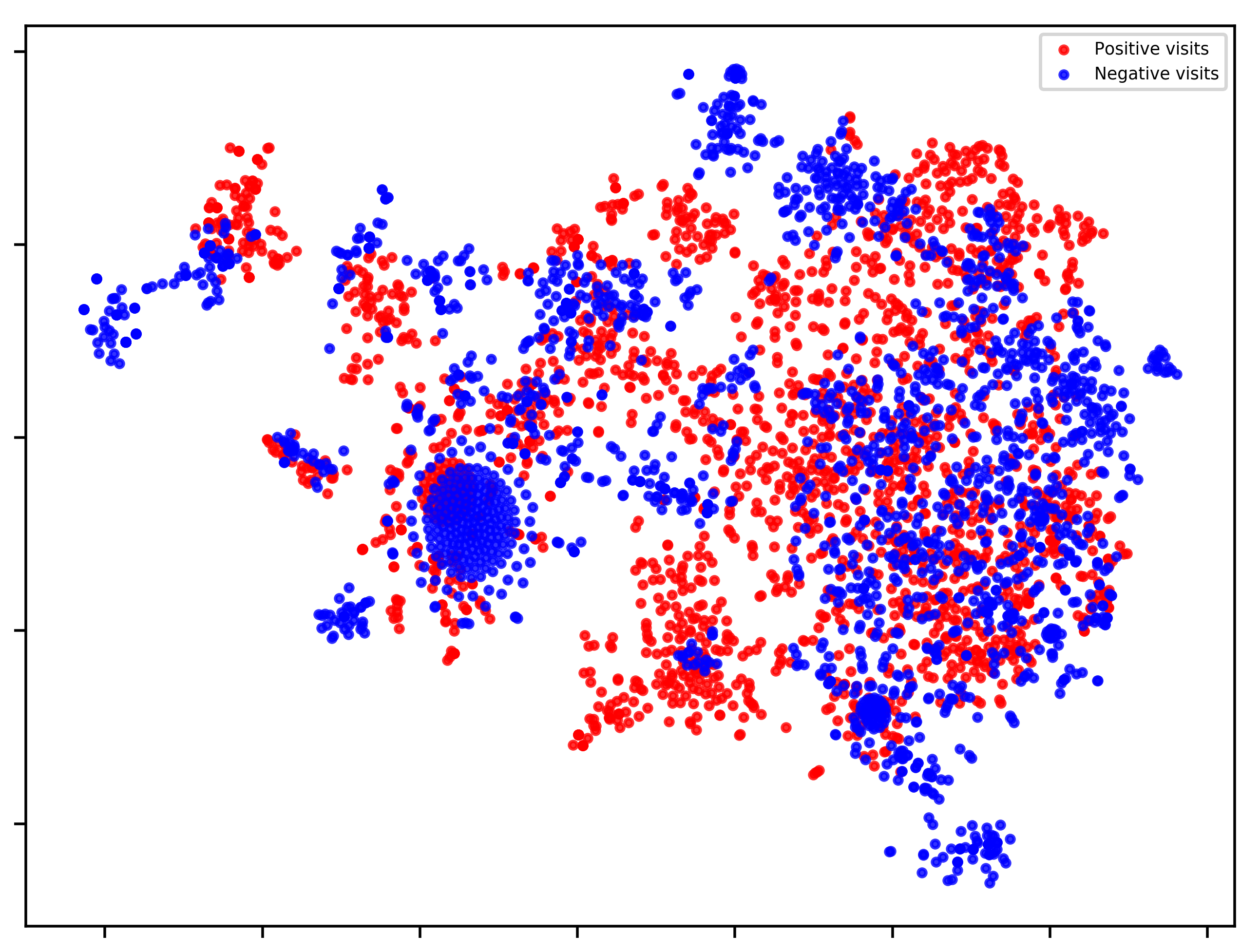}}
  \subfigure[\textbf{CR} (T2D)]{\includegraphics[scale=0.32]{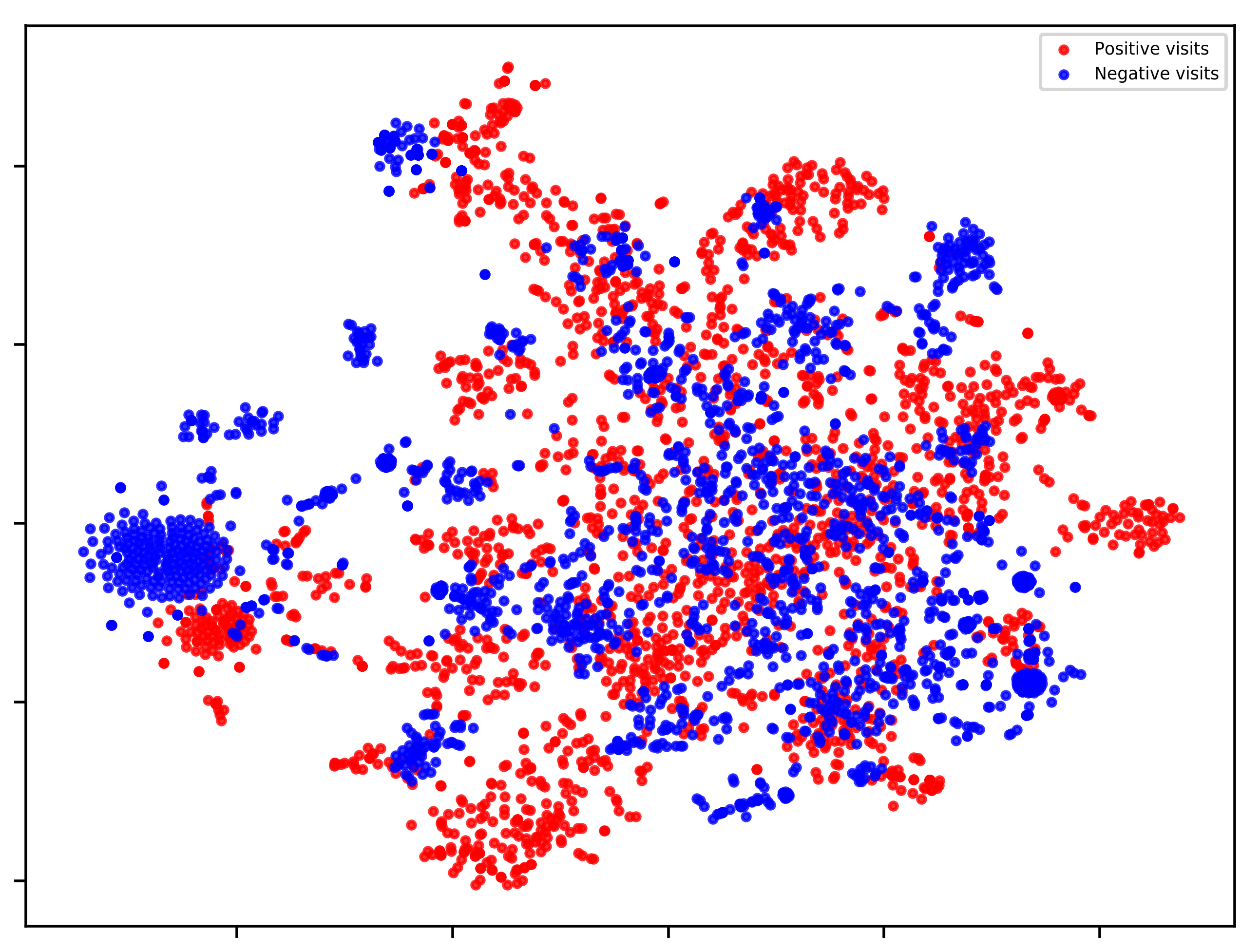}}
  \subfigure[\textbf{CR} + \textbf{VAN} (T2D)]{\includegraphics[scale=0.32]{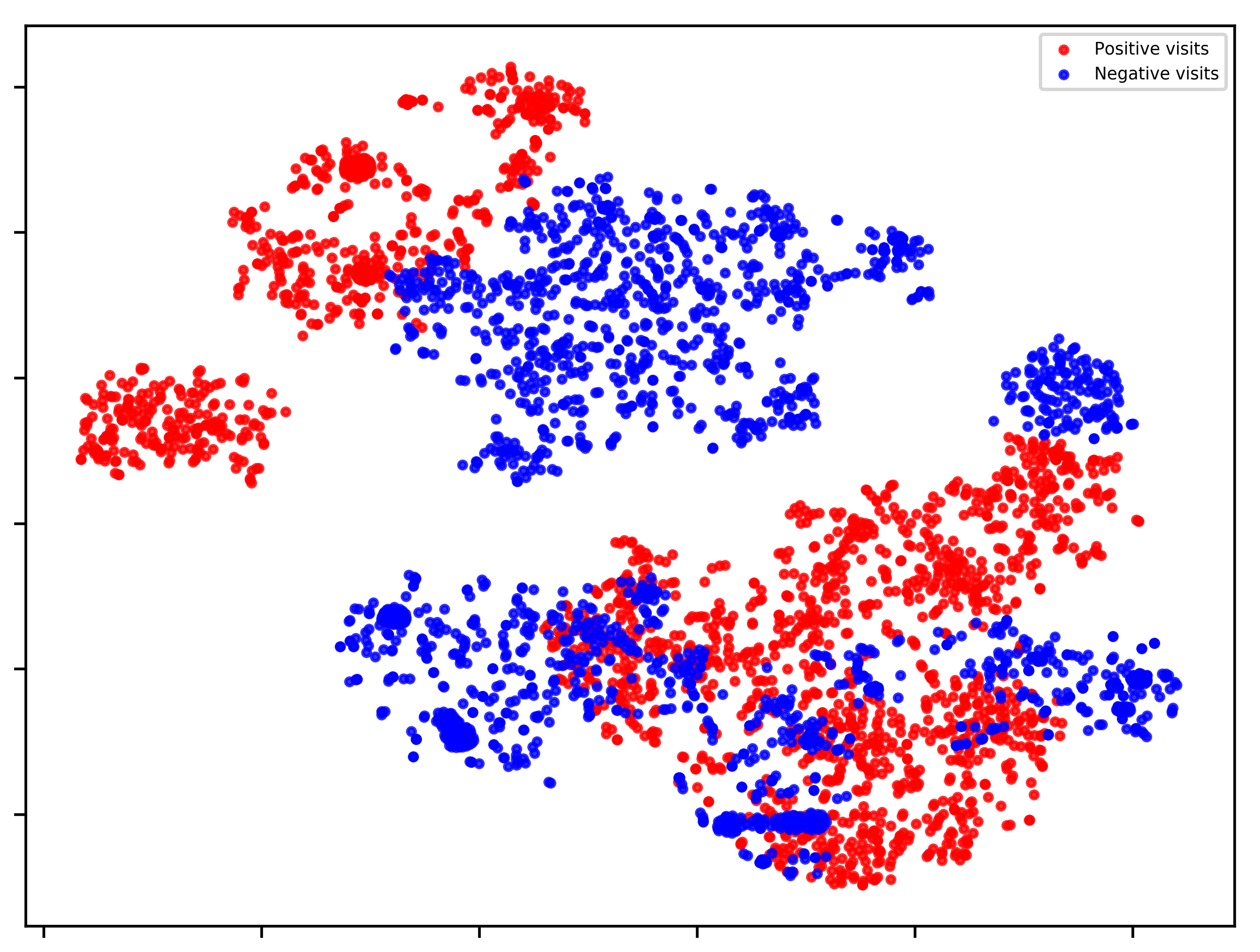}}
    \subfigure[Raw (HF)]{\includegraphics[scale=0.3]{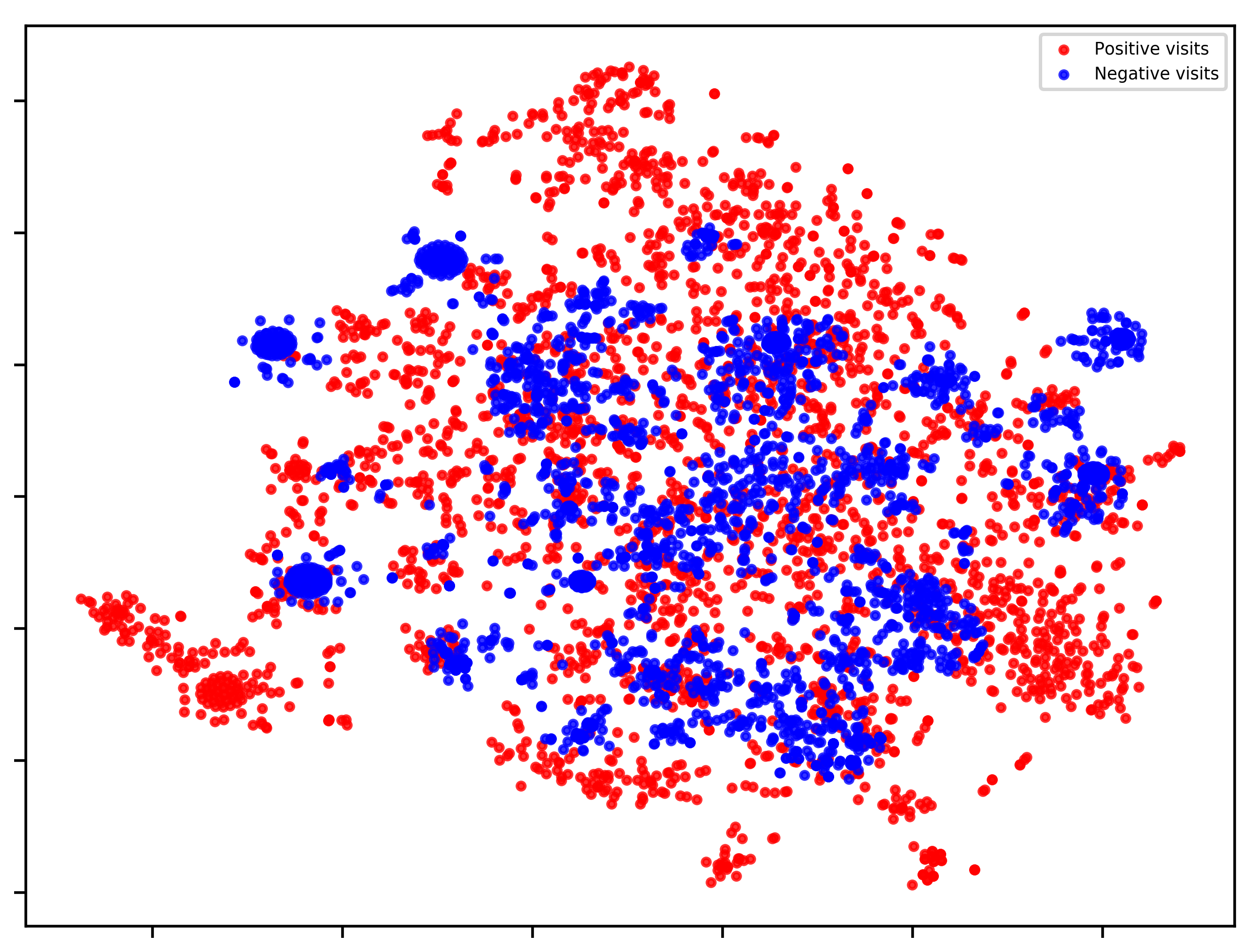}}
  \subfigure[\textbf{CR} (HF)]{\includegraphics[scale=0.32]{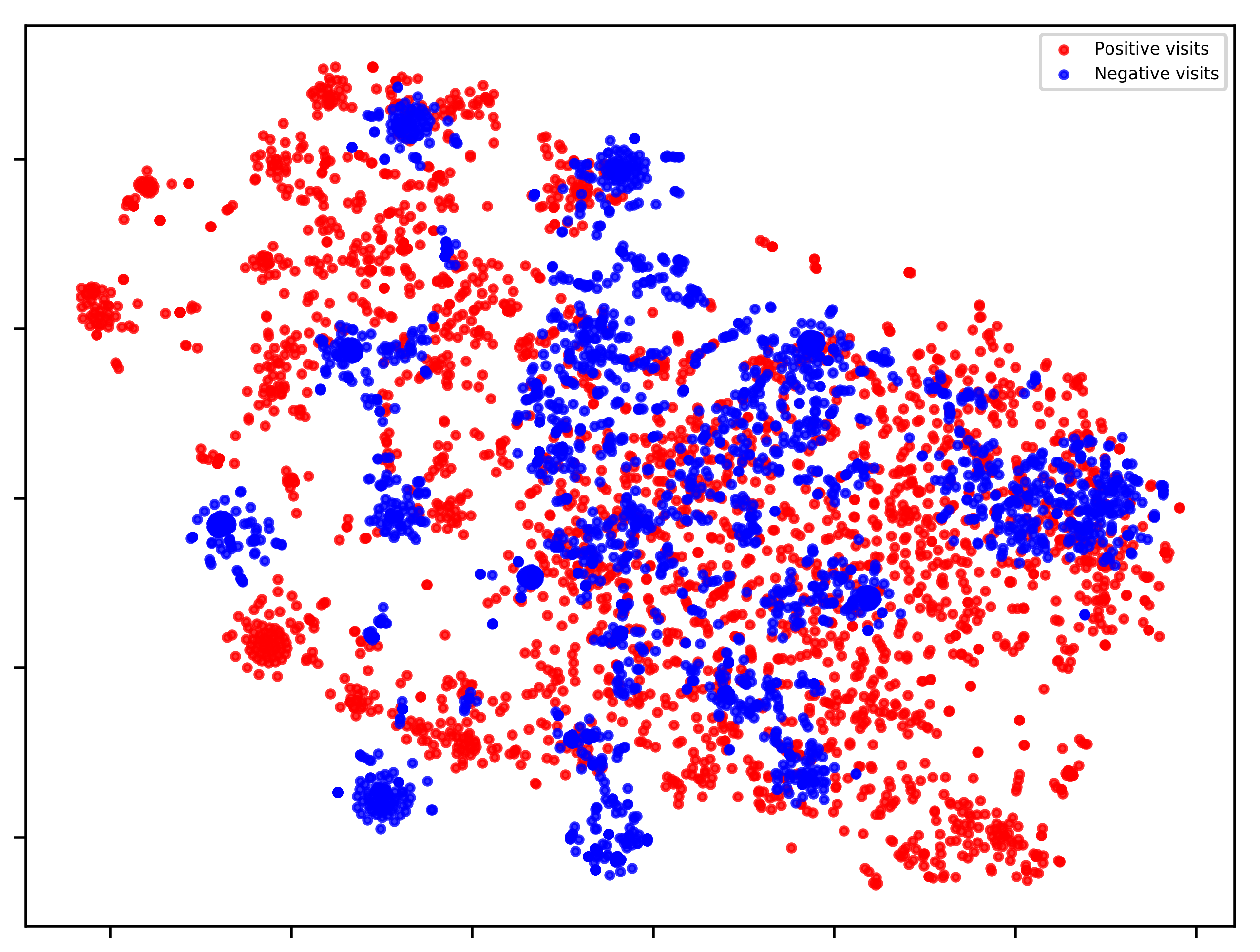}}
  \subfigure[\textbf{CR} + \textbf{VAN} (HF)]{\includegraphics[scale=0.3]{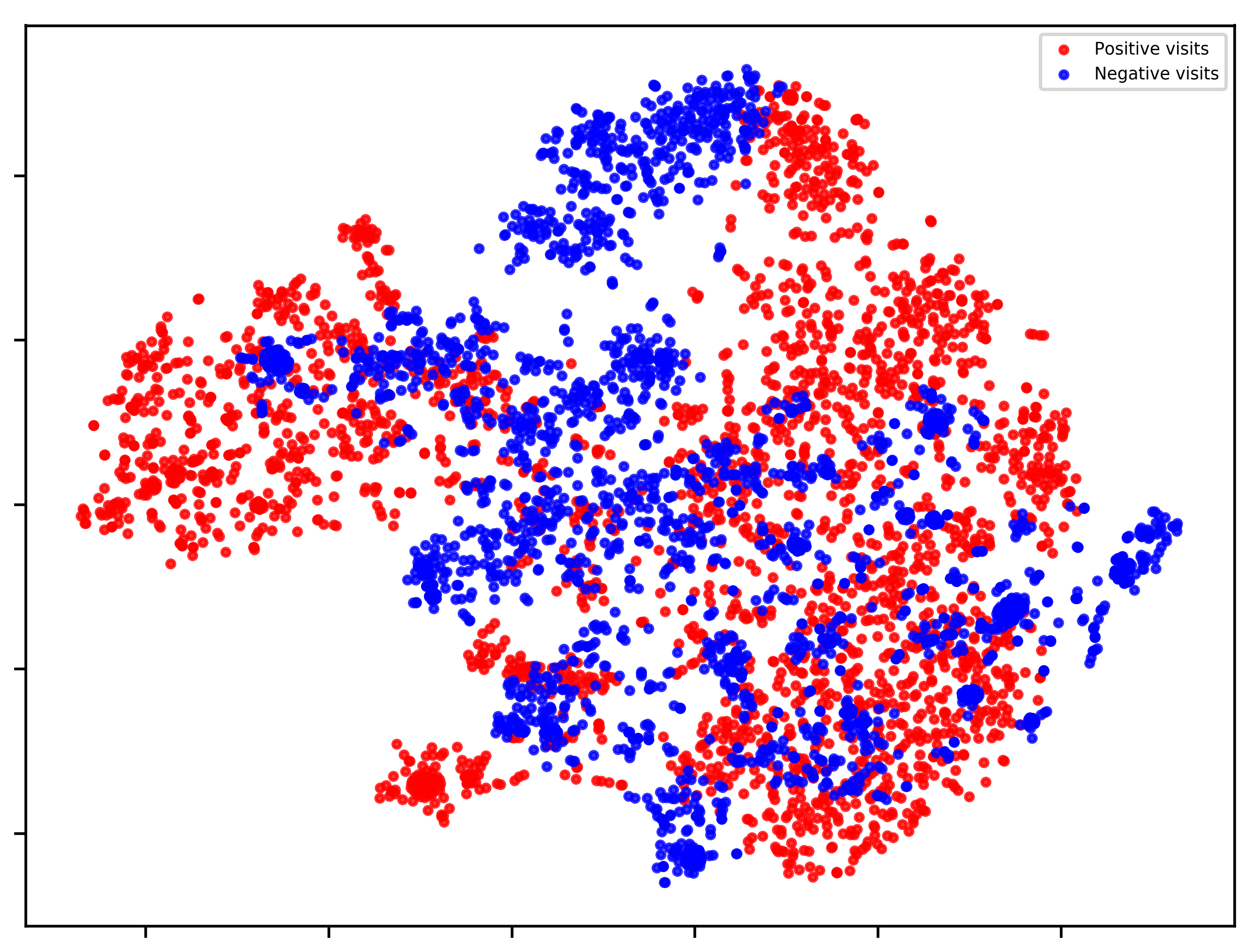}}
  \caption{Visualization of visit embedding vectors from the datasets of type-2 diabetes (T2D) and heart failures (HF) (blue: visits before phenotype onset; red: visits after phenotype onset).}
  \label{fig:visual_visit}
\end{figure*}

\subsubsection{Separation of Case/Control Visits}
We provide visualization of embedding vectors of case/control visits from patients with type-2 diabetes or heart failure in order to show the benefits of the proposed concept re-weighting, visit attention, and the necessity of sequential modeling.  As shown in Figure \ref{fig:visual_visit}, we visualize the raw visit embeddings as in SAMGEP \cite{ahuja2021samgep}, visit embedding with concept re-weighting, and visit embedding with simultaneous concept re-weighting and visit attention.
The case visits, namely after phenotype onset, and control visits, namely before phenotype onset, entangle on the raw embedding space for both phenotypes.
Through concept re-weighting enhancements, the case/control visits get more distinguishable in the embedding space.
By further incorporating the visit attention, the embedding vectors of case visits become further distinguished from control visits. The case/control visits of T2D are well separated in the embedding space even without the additional sequential modeling. It indicates that developing T2D tends to distinctly reshape the patient's clinical status. In this case, a simple classifier such as logistic regression would be able to well identify the onset timings. For HF, though becoming more separated via visit attention, case/control visits are still entangled by large. This indicates the necessity of modeling visit sequential dependency to distinguish the cases from controls for precise incident localization.

\begin{figure*}[tb]
  \centering
  \subfigure[HF onset observed]{\includegraphics[width=0.49\textwidth]{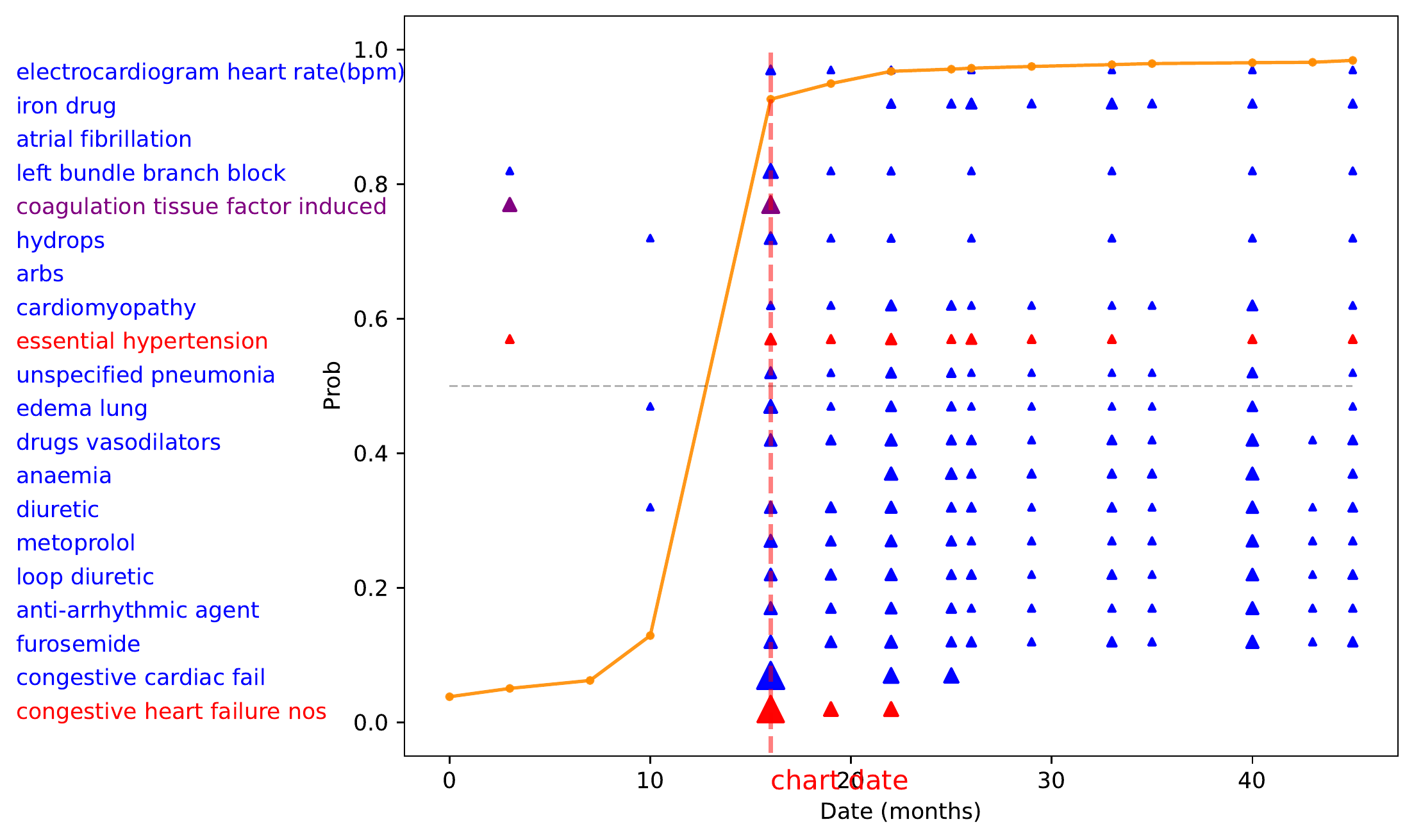}}
  \subfigure[HF onset unobserved]{\includegraphics[width=0.49\textwidth]{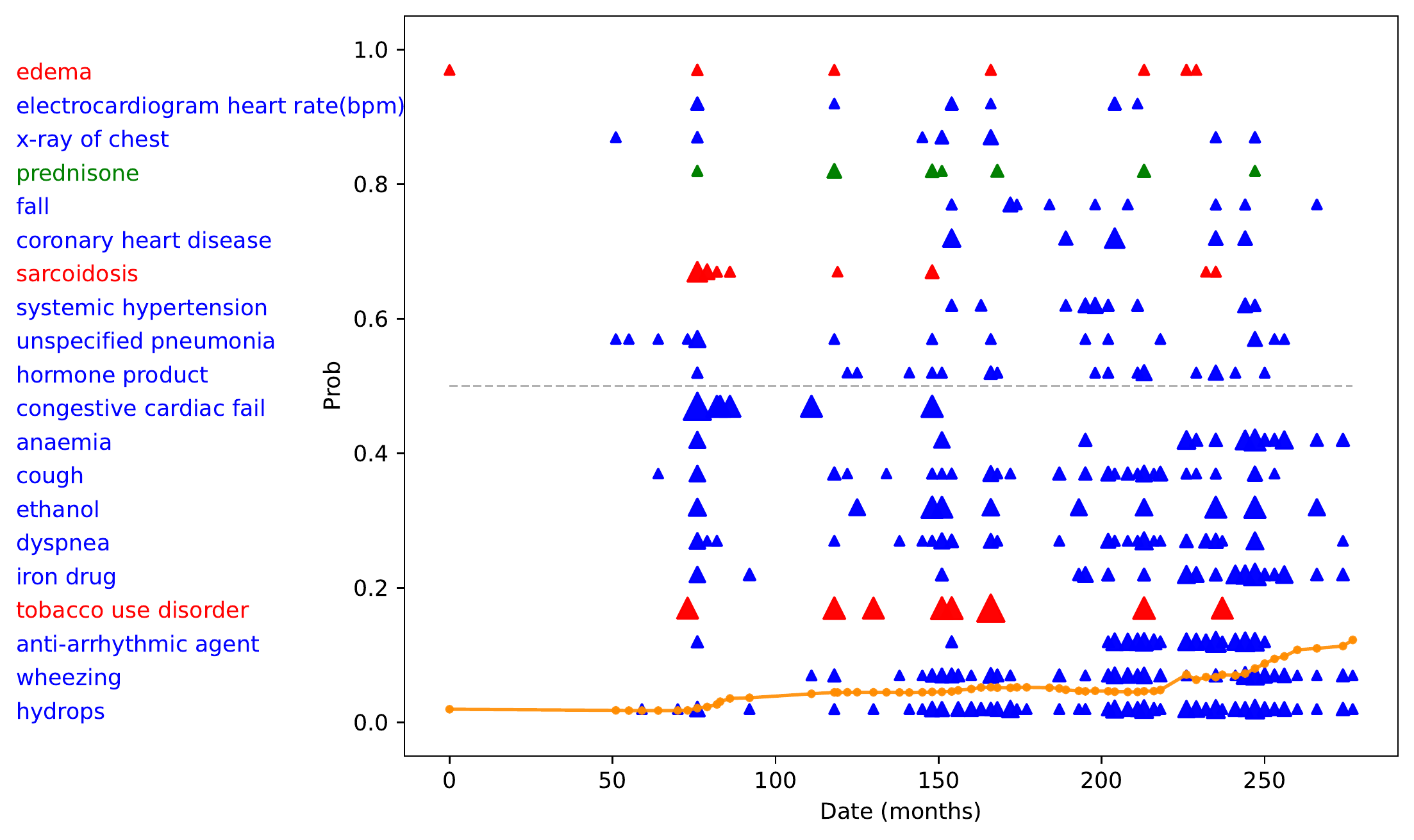}}
  \caption{Visualization of prediction curve (orange) and longitudinal evidence provided by LATTE to help interpretation (red: diagnosis code; green: medication code; purple: lab test code; blue: UMLS CUIs extracted from medical notes; the chart date is the annotated date of HF onset). Concept importance is ranked from bottom to top.}
  \label{fig:visual_interpretation}
\end{figure*}

\subsubsection{Evidence for Prediction Interpretation}
LATTE provides high interpretation capacity by indicating which healthcare concepts at which visits contribute to incident prediction.
As shown in Figure  \ref{fig:visual_interpretation}, we visualize the ``evidence'' at each visit that drives the onset prediction of heart failure. Here, the longitudinal ``evidence'' value is the multiplication of longitudinal concept observations, concept weights, and visit attention values. Each patient could have different longitudinal evidence and the top 20 concepts are visualized.
In Figure \ref{fig:visual_interpretation} (a), the patient's first HF onset is recorded by the EHR data and the provided evidence is highly indicative of heart failure onset. Specifically, LATTE localizes the HF onset at the fifth visit and the top three shreds of evidence provided are ``congestive heart failure nos''(PheCode:428.1), ``congestive cardiac fail''(C0018802), and ``furosemide'' (C0016860) which is a medication used to treat fluid build-up due to heart failure. In Figure \ref{fig:visual_interpretation} (b), the patient is not observed HF onset over the available visits and the top selected concepts are mostly irrelevant to heart failure. For example, the top three pieces of evidence provided by LATTE are ``hydrops''(C0013604), ``wheezing'' (C0043144), and ``anti-arrhythmic agent''(C0003195).

\begin{table}[tp]
  \centering
  \caption{Numbers of risk factors of heart failure among patients with RA from the MGB RA cohort identified by LATTE with the different false positive rate (FPR) and rule-based method. Univariate and multivariate Cox model analyses have been considered for data up to 5 years and 10 years after RA diagnosis.
  }
\begin{tabular}{c|rrrrrrr}
\toprule[1pt]
& \multicolumn{2}{c}{Univariate}
& \multicolumn{2}{c}{Multivariate}\\
    Method & 10 Yrs  & 5 Yrs & 10 Yrs   & 5 Yrs \\
\hline
HF Code & 16 &18 & 19&19\\
% LATTE FPR .10& 16 &18 & 19&19\\
LATTE FPR .05 & 17 &18 & 20&19\\
LATTE FPR .01  & 18 &19 & 20&20\\
\toprule[1pt]
\end{tabular}
\label{Table:survial_risk_factor_summary}
\end{table}

\begin{figure}[tp]
  \centering
  \caption{Estimated hazard ratios with 95 \% confidence intervals for HR risk prediction among RA patients up to 5-year follow-up.}
\label{fig:RAHF_HR}
\begin{center}
\includegraphics[width = 0.7\textwidth]{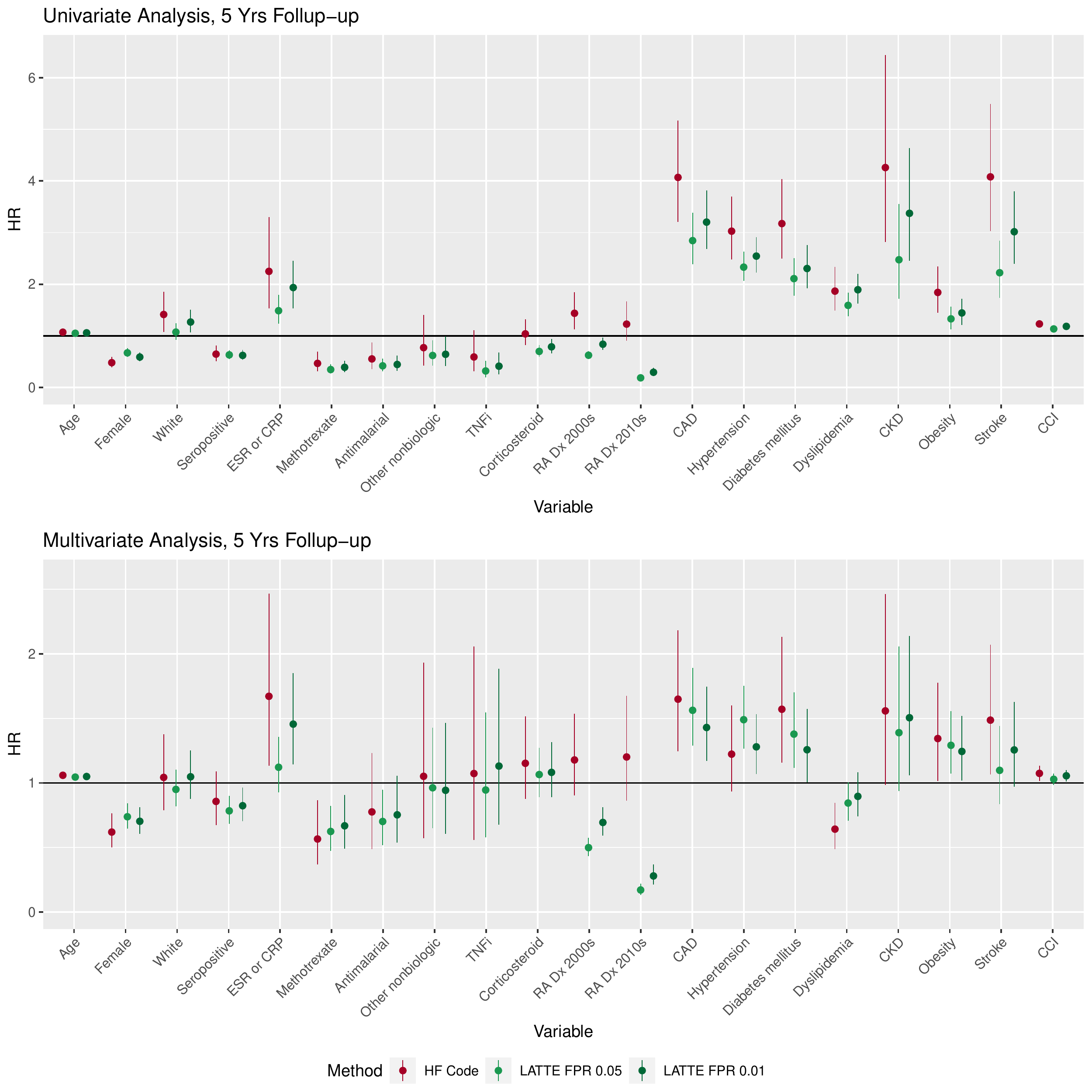}
\end{center}
\end{figure}

\begin{figure}[tp]
  \centering
  \caption{Estimated relative efficiency of analyses with LATTE derived outcomes versus analysis with HF diagnosis code derived outcomes in HR risk prediction among RA patients up to 5-year follow-up. LATTE-derived outcomes systematically improved the efficiency. }
\label{fig:RAHF_RE}
\begin{center}
\includegraphics[width = 0.7\textwidth]{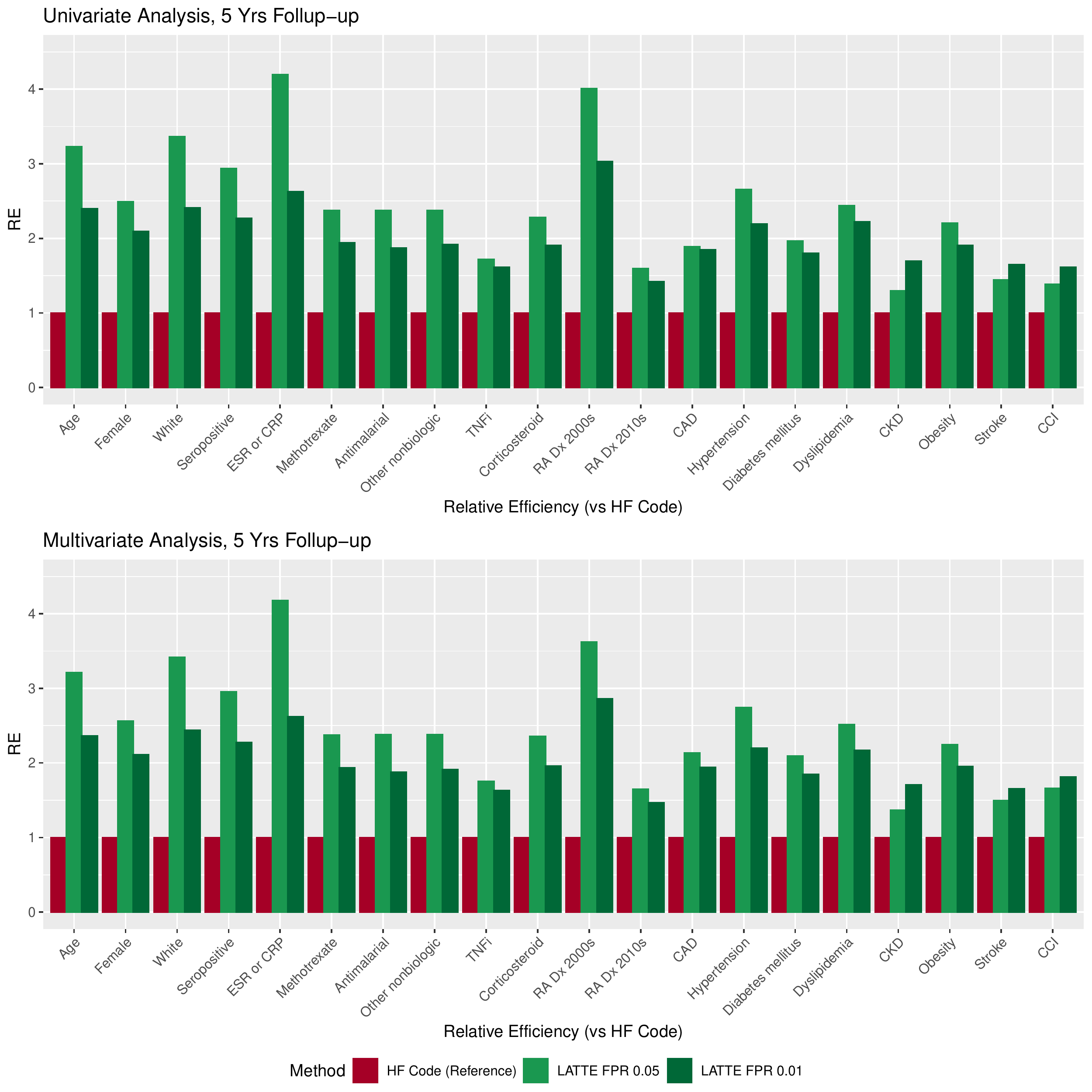}
\end{center}
\end{figure}

\subsection{Incident Phenotyping for Identifying Risk Factors of Heart Failure}
Heart failure is a significant cause of morbidity and mortality in patients with RA \cite{nicola2005risk}.
We revisit the HF study among the MGB RA cohort described in Section \ref{sec_experiment_setting} on the risk factors associated with heart failure among patients with rheumatoid arthritis \cite{huang2021association}.
The covariates considered include demographics, recent lab test results, prior medications, calendar time of RA diagnosis, comorbidities, and cardiovascular disease history before RA diagnosis (details on the definition and extraction of the covariates provided in the original paper). Among the 9087 RA patients in the study, 1219 (13.2\%) have at least one HF diagnosis code in their EHR after RA diagnosis. While the lack of HF code almost consistently indicates free from HF, the presence of HF code contains many false positives. Exact HF status and timings for 102 patients sampled from the subset with HF diagnosis code are annotated from chart review, among which 33 (32.3\%) have evidence of HF within 10 years from RA diagnosis. The 10-year prevalence of HF is thus 4.5\%. In the original analysis by \cite{huang2021association}, the time of the first HF code after RA diagnosis is used to define the HF time. Patients with no HF code are marked as censored at the last EHR encounter date.  Four sets of analyses are employed to assess the risk factors. Two maximal follow-up times, 10 years or 5 years after RA diagnosis, are considered. Patients at risk at the maximal follow-up, defined as not having HF code nor reaching the last EHR encounter, are censored at the maximal follow-up. Two regression methods are considered, a) a univariate Cox model that separately regresses the risk factors with the time-to-HF-code, and b) a multivariate Cox model that regress all risk factors with the time-to-HF-code.

Using the LATTE prediction on longitudinal HF incidence rate, we substitute the time-to-HF-code with the LATTE-derived HF onset time. According to a threshold to be described later, we define the LATTE-derived HF onset time as the first time the longitudinal HF incidence rate from LATTE exceeds the chosen threshold. We select the threshold according to the false positive rate (FPR) for HF status at the last EHR encounter and evaluate all patients without HF code and 102 labeled patients up weighted by the inverse labeling probability among patients with HF code (1/0.084). To investigate the impact of the threshold, we consider 2 thresholds targeting .05, and .01 FPR.

In Table \ref{Table:survial_risk_factor_summary}, we present the risk factor detection results using the time-to-HF-code outcome and LATTE-derived HF time with 2 thresholds. The numbers in the table are the counts of risk factors whose 95\% confidence intervals of relative risks do not contain one.
By tightening the tolerance of FPR from .05 to 0.01, analysis with LATTE-derived HF time starts to detect more potential risk factors. Considering lower FPR means a larger threshold and subsequently a smaller number of derived HF events, we suggest that LATTE at 0.01 FPR might have filtered out many spurious HF events and hence eliminate the bias toward null induced by them. In Figure \ref{fig:RAHF_HR}, we present the point estimates along with 95\% confidence intervals up to 5-year follow-up.
In Figure \ref{fig:RAHF_RE}, we present the estimated relative efficiency of coefficient estimation from LATTE-derived HF times in comparison with that from the time-to-HF-code outcome up to 5-year follow-up, which demonstrates a systematic advantage for  LATTE-derived HF times.
We present the results for the analyses up to 10-year follow-up
in the Supplementary Note 1 which similarly shows the advantages of LATTE.
Another notable discovery from the analyses with LATTE-derived HF time is the decreasing trend of HF risk over calendar time. In the LATTE at .01 FPR analysis, patients diagnosed with RA after 2000 are associated with 55\% risk reduction in univariate analysis and 61\% risk reduction in the multivariate analysis compared to patients diagnosed before 2000, and patients diagnosed with RA after 2010 are associated with 85\% risk reduction in both analyses compared to patients diagnosed before 2000.
The finding may suggest the progress in managing cardiovascular health among patients with RA during the past decades. Such temporal trending on health outcomes has also been reported for other EHR-based studies with long observation windows \cite{hou2022temporal}.

\section{Method}
\subsection{Overview}
\begin{figure*}[]
\begin{center}
\includegraphics[width= \textwidth]{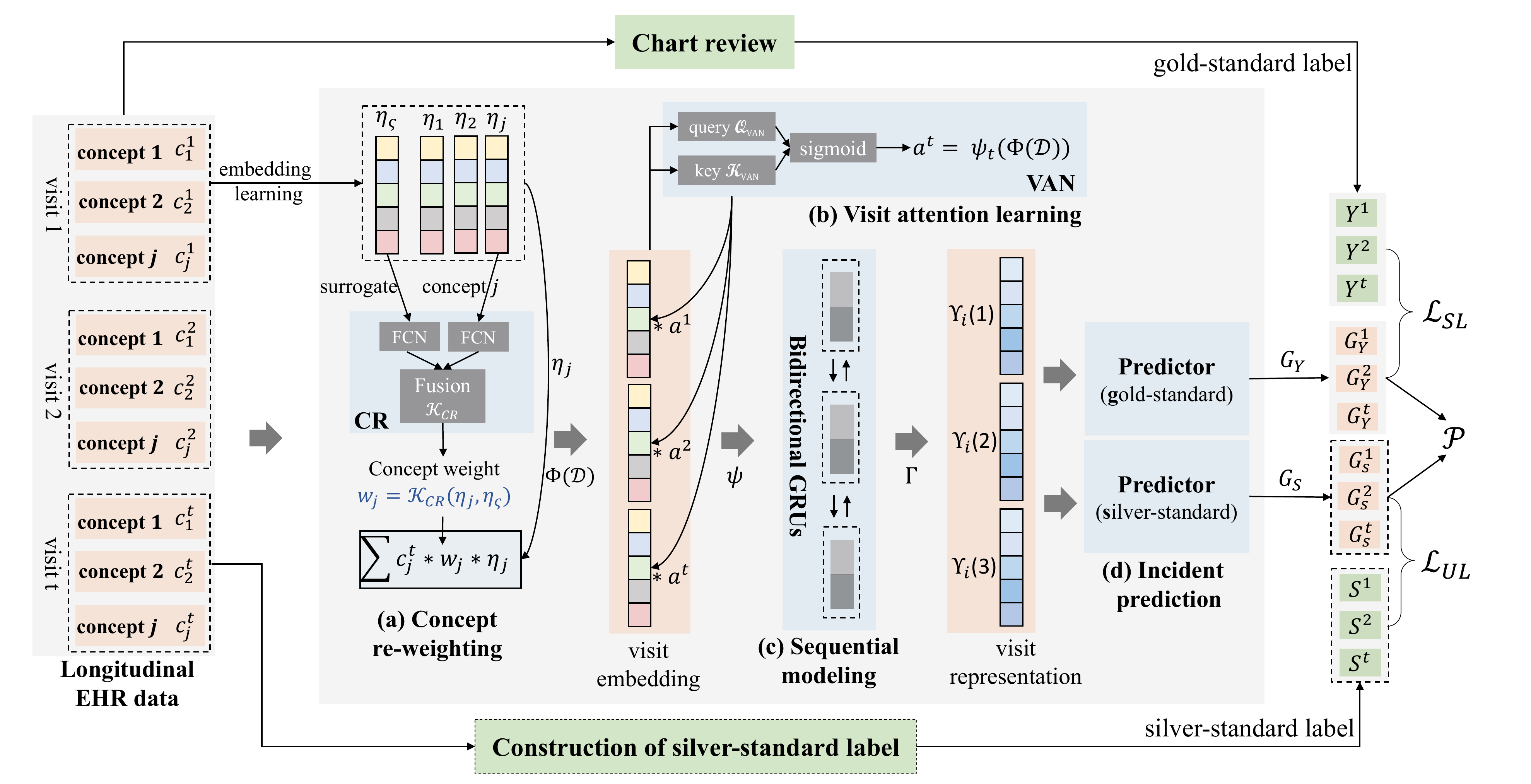}
   \caption{Architecture of LATTE which is jointly trained by the gold-standard labels and silver-standard labels. The computation of LATTEs consists of concept re-weighting, visit attention learning, visit temporal modeling and incident prediction.}
   \label{figure:pipeline}
\end{center}
\end{figure*}

\textbf{Architecture of LATTE}: As illustrated in Figure \ref{figure:pipeline},  with longitudinal EHR data and a small number of labels on the phenotype status over time as the input, LATTE consists of four key computational components: (a) a \textbf{C}oncept \textbf{R}e-weighting (\textbf{CR}) module that assigns a weight for each input concept or feature; (b) a \textbf{V}isit \textbf{A}ttention \textbf{N}etwork (\textbf{VAN}) which aims to assign higher weights to visits that are more indicative of the incident; (c) a sequential model to capture visit temporal dependency and obtain visit representations; (d) final incident predictions at each visit by the \textbf{P}redictor.

\textbf{Procedures of LATTE}: To alleviate the need for gold-standard labels, the learning of LATTE comprises the following two steps. (i) LATTE constructs longitudinal silver-standard labels for the event status over time based on predictive surrogates to perform unsupervised model pre-training. (ii) LATTE is fine-tuned jointly by the gold-standard labels and silver-standard labels. The four computational components of LATTE are optimized end-to-end together at both the pre-training and fine-tuning stages.

% We first introduce the input data structure and notations in Section \ref{DLdata}, including the preparation of semantic embedding for input features or concepts in Section \ref{embeding} and the construction of silver standard labels in Section \ref{silver}. Then, we illustrate the model construction and training process of LATTE in Section \ref{model+train}, which includes the model's architecture, computation flow in Section \ref{DLmodel}, loss function in Section \ref{DLloss} and label-efficient training strategy in Section \ref{sec_Label_efficient_learning}. Finally, we also introduce a contrastive training objective to improve the cross-site portability of LATTE in Section \ref{sec_portability}.

\subsection{Input Data Structure and Notations}\label{DLdata}
EHR data consists of longitudinal patient visits, with each visit recording the observations of both structured EHR concepts (codes for diagnosis, procedures, medication prescriptions, laboratory tests orders, and results), and unstructured narrative clinical notes.  Raw EHR concepts are rolled up to higher level concepts according to common ontology as in previous studies \cite{hong2021clinical}: Diagnosis (ICD) codes are grouped into the established Phenome-wide association studies (PheWAS) catalogs (PheCodes); Procedure codes are grouped into categories according to the Clinical Classifications Software for Services and Procedures (CCS), medications are mapped to ingredient level RxNorm codes, and laboratory tests are mapped to Logical Observation Identifiers Names and Codes (LOINC). From free-text clinical notes, we extract NLP mentions of clinical terms mapped to the Concept Unique Identifiers (CUIs) in the Unified Medical Language System (UMLS) using the Narrative Information Linear Extraction (NILE) tool \cite{yu2013nile}.

We assume that the training data contains a total of $N$ patients, organized in a longitudinal format indexed by $t$ and that without generality the first $n$ patients are labeled.
For patient $i$ and visit at time $t \in \{1,\dots,T_i\}$,
let $Y_i^t \in \{0,1\}$ denote the gold-standard label on the event status, which is annotated by physicians,
$S_i^t \in [0,1]$ be the silver-standard label for incidence status,
and $\bD_i^t \in \R^p$ be the $p$-dimensional vector of features, where
$T_i$ is the last follow-up time for patient $i$,
upper bounded by global maximal follow-up time $t_{\max}$. Here, the scalable silver-standard labels are obtained in an unsupervised manner.
They are expected to be indicative of gold-standard labels but may be contaminated by bias or noise. For learning the onset time of a phenotype, silver-standard labels are typically derived from the corresponding diagnosis codes or NLP mentions.
Hence, the input data consists of
$\Lscr=\{(Y_i^t,S_i^t, \bD_i^t): t=1,\dots,T_i, \, i=1,\dots,n\}$ for $n$ labeled patients
and $\Uscr = \{(S_i^t, \bD_i^t): t=1,\dots,T_i,\,i=n+1,\dots,N\}$ for the other $N-n$ unlabeled patients.
Let $\Dscr_i = \{\bD_i^t: t=1,\dots, T_i\}$
be the features aggregated over follow-up time.
Our LATTE algorithm builds a prediction model for $\Pbb(Y_i^t = 1\mid \Dscr_i)$, $t=1,\dots,T_i$, the
incidence rate over time given the longitudinal features.

\subsubsection{Semantic Embedding Vector of Input Features}\label{embeding}
LATTE leverages a $q$-dimensional semantic embedding vector as prior knowledge of each element of the $p$-dimensional feature $\bD$
to compress visit information into
a $q$-dimensional visit embedding $\phi(\bD_i^t)$.
Let $\bge_1,\dots,\bge_p \in \R^q$ be the semantic embedding vectors associated with features or concepts in $\bD$ and $\bge_{\varsigma}$ be the semantic embedding vector of the surrogate concept which can be the target phenotype itself or a concept that is most predictive of it.
The embedding vectors are obtained by performing matrix factorization, a variant skip-gram algorithm \cite{mikolov2013distributed,levy2014neural} on a co-occurrence matrix of the EHR concepts aggregated from large-scale unlabeled patient-level longitudinal EHR data \cite{hong2021clinical}.
As shown in \cite{beam2019clinical,hong2021clinical,zhou2022multiview}, such embedding vectors effectively capture the clinical semantic relationship or similarity of EHR concepts or input features.

\subsubsection{Construction of Longitudinal Silver Labels}\label{silver}
The silver standard label $S_i^t$, which serves as a noisy proxy for $Y_i^t$, can be designed according to specific applications. When the event information can be well captured by surrogate features such as a relevant NLP CUI or PheCode,
LATTE constructs longitudinal silver-standard labels by leveraging such surrogate concepts similarly to other weakly supervised algorithms such as PheNorm \cite{yu2018enabling}. Let $c_i^t$ denote the counts of silver-standard label concepts at visit $t$ and $U_i^t$ be a health utilization measure at $t$ which is often needed for normalizing the silver standard labels.

When the outcome of interest is time to the first onset of a condition
with $Y_i^t$ indicating whether the event has occurred by time t, we use the cumulative counts $\sum_{u=0}^t c_i^u$ up to visit $t$,
\begin{equation}
S_{\CUM,i}^t=\expit\left[\left\{\log\left(1+\sum_{u=0}^{t} c_i^u\right)-\alpha \log\left(1+\sum_{u=0}^{t} U_i^u\right)\right\}/\tau\right]
\label{eq:silver_label_cum}
\end{equation}
For prediction of recurrent events such as relapse status over time, with $Y_i^t$ episodically shifting between $0$ and $1$, we only use $c_i^t$ at visit $t$,
\begin{equation}
S_{\REC,i}^t=\expit[\{\log(1+c_i^t)-\alpha \log(1+U_i^t)\}/\tau].
\label{eq:silver_label_rec}
\end{equation}
\noindent The hyper-parameter $\tau$ denotes the temperature of the $\expit(\cdot)$ and controls the sharpness of the silver-standard label. A small $\tau$ would sharpen the silver labels to be in alignment with the gold-standard label which is 0 for a negative visit and 1 for a positive visit. $\alpha$ controls the influence of $U_i^t$ for constructing silver-standard labels

\begin{table}[h]
  \centering
  \small
  \caption{Summary of the used notations.
  }
    \begin{tabular}{cccc}
    \toprule[1pt]
    Notation&Description& Notation&Description \\
    \midrule
    $T_i$ &  last follow-up time for patient $i$ &
    $t_{\max}$ & global maximal follow-up time\\
    $Y_i^t$ & incidence status for patient $i$ at visit $t$ &
    $S_i^t$ & silver-standard label for patient $i$ at visit $t$
    \\
    $\bY_i$ & aggregation of $Y_i^t$ for patient $i$ across visits &
    $\bS_i$ & aggregation of $S_i^t$ for patient $i$ across visits \\
    $\bD_i^t$ & features for patient $i$ at visit $t$ &
    $\Dscr_i$ & aggregation of $\bD_i^t$ for patient $i$ across visits\\
    $\bF$ & visit representation module &
    $\Upsilon_i$ & visit representation of patient $i$\\
     $c_i^{t}$    & count of concept $i$ at visit $t$        & $\bge$&semantic concept embedding \\
%    $c_i^{t}$   & count of concept $i$ at visit $t$        &  $U_i$   &    health utilization measure at $t$ \\
%    $w_{Y,i}^t$& incidence kernel-weight of patient $i$ at visit $t$&$w_{S,i}^t$&silver kernel weight of patient $i$ at visit $t$ \\
%        $\alpha$ &   weight of health utilization in silver label   &  $\tau$   &    temperature of silver label\\
        $\mathcal{L} $             & training loss        &
 $\Pcal$   &    penalty terms\\
       $\bgG$& Bi-GRU deep-learning module      &  N  &   number of training patient\\

        \textbf{CR}             & concept re-weighting module        &   \textbf{VAN}       &    visit attention network\\
        $\bgf$& concept re-weighting transformation &$\bgps$& visit attention transformation\\
      $\Kcal_{\CR}$&parameter of concept re-weighting&$\bgTh$&parameters of all layers\\
      $\bQcal_{\VAN}$& query matrix of VAN &$\bKcal_{\VAN}$&key matrix of VAN\\
      %$\textbf{P}_{gold}$             & predictor of gold-standard label     &    $\textbf{P}_{silver}$             & predictor of silver label \\
      $G_Y$ & logistic transformation for incidence & $G_S$ & logistic transformation for silver labels\\

%      $\bgb_Y$ &parameter of $G_Y$&$\bgb_S$&parameter of $G_S$\\

%    $\lambda$ & weight of penalty terms & $m_c$  & threshold for contrastive learning  \\
%      $\gamma $     &  weight of silver-label loss\ & $\kappa$& weight of cross-site loss \\
      $\widetilde{\Lscr}$& generated synthetic data & $\widetilde{A}_{i,k}$& time window of synthetic data\\
 %     $\widetilde{Y}_{i,k}$& incidence of synthetic data &$\widetilde{Y}_{i,k}^{(1)},\widetilde{Y}_{i,k}^{(2)}$& mid product of generating incidence\\
%      $\widetilde{T}_{i,k}$& follow-up time of synthetic data &$\widetilde{T}_{i,k}^{(1)},\widetilde{T}_{i,k}^{(2)}$& mid product of generating follow-up time\\
%      $\widetilde{\bD}_{i,k}$& features of synthetic data &$\widetilde{\bD}_{i,k}^{(1)},\widetilde{\bD}_{i,k}^{(2)}$& mid product of generating features\\
%    $L_{i,k}$& start-time of synthetic sub-sequences&$ C_{i,k}$& end-time of synthetic sub-sequences\\

    \bottomrule[1pt]
    \end{tabular}%
    \label{tab:notations}
\end{table}

\subsection{Architecture and Training of LATTE}\label{model+train}
% \subsection{Computational Flow and Training}\label{model+train}

We next describe the construction of deep-learning models
$G_Y\circ \bF(t,\Dscr_i)$ for incidence $\Pbb(Y^t_i=1\mid \Dscr_i)$
and $G_S\circ \bF(t,\Dscr_i)$ for silver-standard label $\Ebb(S^t_i\mid \Dscr_i)$.
Inspired by semi-supervised learning under a semi-parametric transformation model \cite{HouChanWangCaiBx21},
we let the models for incidence $Y_i^t$ and silver-standard label $S_i^t$ share the
core visit representation learning component $\bF(t,\Dscr_i) \in \R^q$ while allowing different prediction functions $G_Y$ and $G_S$: $\R^q \mapsto [0,1]$.
We create in $\bF(t,\Dscr_i)$: 1) a Concept Re-weighting module
to learn incident-indicative input concepts; 2) a Visit Attention Network to highlight informative visits from background noises among other visits; and 3) the bi-directional Gated Recurrent Unit (Bi-GRU) network for communication over time.
In Section \ref{DLloss}, we describe the loss functions
for the training of deep-learning model $F$.
Based on the cross-entropy loss of the binary outcome $Y_i^t$, we devise 1) a kernel weighting strategy
to borrow outcome data near $t$ for learning $G_Y\circ \bF(t,\Dscr_i)$;
2) a penalty to regularize $G_Y\circ \bF(t,\Dscr_i)$ toward a function
increasing/smooth along $t$ and compatible with
the context of the outcome; and
3) a semi-supervised training strategy to leverage the
informative silver-standard labels $S_i^t$.

\subsubsection{Construction of Deep-learning Model}\label{DLmodel}

As illustrated by the model architecture in Figure: \ref{figure:pipeline}, the incident prediction is produced through the following steps:
\begin{enumerate}
    \item Input layer: longitudinal $\Dscr_i = \{\bD_i^1,\dots,\bD_i^{T_i}\}$;
    \item Concept re-weighting module: mining the input concepts' semantic relationship to the target phenotype,
    %dimension reduction of feature space,
    $$\bgf: \bD_i^t\in\R^p \mapsto \bgf(\bD_i^t) \in \R^q, \, q\ll p,\,
    \bgF(\Dscr_i) = (\bgf(\bD_i^1), \dots, \bgf(\bD_i^{T_i})).
    $$
    \item Visit attention module:
    obtaining the visit attentions by contrasting $\phi(\bD_i^t)$ along
$t$,
    $$\bgps: \bgF(\Dscr_i)\in \R^{q\times T_i} \mapsto \bgps\circ \bgF(\Dscr_i) \in \R^{T_i}, \,
    \bgps = (\psi_1,\dots, \psi_{T_i})^\top.
    $$
    \item Sequential modeling module: communicating the visit embeddings along time through a Bi-GRU layer:
    $$
    \bgG: \{\bgF(\Dscr_i), \bgps(\Dscr_i)\} \mapsto
     \{\bF(1,\Dscr_i),\dots,\bF(T_i,\Dscr_i)\}, \,
     \bF(t,\Dscr_i) \in \R^q.
    $$
    \item Incident predictor:
    separating the logistic regression transformations for incidence and silver-standard label models with link
    $\expit(x) = 1/(1+e^{-x})$,
    \begin{align*}
    &\text{for incidence }
    G_Y\{\bF(t,\Dscr_i)\} = \expit\{\beta_{Y,0}+\bgb_Y^\top\bF(t,\Dscr_i)\}, \\
    &\text{for silver-standard label }
    G_S\{\bF(t,\Dscr_i)\} = \expit\{\beta_{S,0}+\bgb_S^\top\bF(t,\Dscr_i)\}.
    \end{align*}
\end{enumerate}
We detail the design of $\bgf$ with parameter $\Kcal_{\CR}$, $\bgps$ with parameters $(\bQcal_{\VAN},\bKcal_{\VAN})$
and $\bgG$ in the following paragraphs. The final models are determined by the combined parameters of $\bgf$, $\bgps$,
$\bgG$, $G_Y$ and $G_S$ across all layers, namely
$$
\bgTh = (\Kcal_{\CR}, \bQcal_{\VAN}, \bKcal_{\VAN}, \bgG,
\beta_{Y,0},\bgb_Y, \beta_{S,0},\bgb_S).
$$

\textbf{Concept Re-weighting: Selecting Important Features.}
LATTE learns a \textbf{C}oncept \textbf{R}e-weighting (\textbf{CR}) module to attach a weight to each input concept by mining its semantic relationship to the target phenotype, which reduces colinearity of features and overfitting along irrelevant features.
We characterize the relevance of
features by a Multi-Layer Perception (MLP) network:
$$
\Kcal_{\CR}: (\bge_j,\bge_{\varsigma}) \in \R^q\times\R^q
\mapsto \Kcal_{\CR}(\bge_j,\bge_{\varsigma})\in \R.
$$
$\Kcal_{\CR}$ has two input branches, such that one branch receives the embedding vector of the input concept and the other one receives the embedding vector of the target phenotype. Both branches first use one fully-connected layer to learn a low-dimensional representation, which is then fused to output the importance of the input concept, ranging from 0 to 1. The $\Kcal_{\CR}$ module is shared across all concepts and optimized end-to-end to learn the weight of each input concept.

Using standardized weights derived from $\Kcal_{\CR}$, we aggregate the concept embedding vectors within each visit to obtain the visit embeddings:
\begin{equation}
\bgf(\bD_i^t;\Kcal_{\CR})=\frac{1}{p}\sum_{j=1}^p D_{i,j}^{t}  \frac{\exp\{\Kcal_{\CR}(\bge_j, \bge_{\varsigma})\}}{\sum_{j'=1}^p \exp\{\Kcal_{\CR}(\bge_{j'}, \bge_{\varsigma})\}} \bge_j,
\label{eq:visit_embedding}
\end{equation}
Such a competing normalization across all $p$ concepts would induce visit embeddings dominated by the most informative features, thus eliminating the noise from irrelevant features.

Compared to the direct utilization of embedding similarity between $\bge_j$ and $\bge_{\varsigma}$ as the concept importance,
our data-driven module $\Kcal_{\CR}$ would
adaptively capture the predictability of features even if the semantic similarity aligns poorly with
predictability of incidence.

\textbf{Visit Attention: Highlighting Informative Visits.}
In the next step, LATTE devises a \textbf{V}isit \textbf{A}ttention \textbf{N}etwork (\textbf{VAN}) to highlight informative visits according to the attention value. The \textbf{VAN} module can reduce noise from non-informative visits adaptively for patients with heterogeneous background noises.
In \textbf{VAN}, we employ a self-attention \cite{vaswani2017attention} layer to contrast informative visits from the whole visit sequence in the background. Self-attention \cite{vaswani2017attention} has shown to be a successful technique in capturing long-range sequential dependency for various data types including texts \cite{devlin2018bert} and videos \cite{arnab2021vivit}.
The network \textbf{VAN} receives sequential visit embedding vectors to obtain corresponding attention values. There are two components in the \textbf{VAN} shared across all visits:
\begin{align*}
&\text{Query } \bQcal_{\VAN}: \phi(\bD_i^t) \in \R^q \mapsto \bQcal_{\VAN}\circ  \phi(\bD_i^t) \in \R^d,\\
&\text{Key } \bKcal_{\VAN}: \phi(\bD_i^t) \in \R^q \mapsto \bKcal_{\VAN}\circ  \phi(\bD_i^t) \in \R^d,
\end{align*}
both of which consist of a linear mapping layer as in \cite{vaswani2017attention} to map the $q$-dimensional visit embedding to the $d$-dimensional query and key vectors, respectively. The attention value for visit $t$ of patients $i$ is derived from averaging the inner products between its query vector and the key vectors across all available $T_i$ visits, including itself,
\begin{equation}
\psi_t ( \bgF(\Dscr_i); \bQcal_{\VAN},\bKcal_{\VAN})= \frac{1}{T_i}\sum_{u=1}^{T_i} \mathrm{expit}\left[\bQcal_{\VAN}\{\bgf(\bD_i^u)\}^\top \bKcal_{\VAN}\{\bgf(\bD_i^t)\}/\sqrt{d}\right],
\label{eq:visit_attention}
\end{equation}
\noindent where $\mathrm{expit}(x)=1/(1+e^{-x})$. We choose the $\mathrm{expit}$ function rather than the typical soft-max function \cite{vaswani2017attention}  because there could be multiple visits or incidents to pay attention to, and not just only sparse visits as encouraged by soft-max.

\textbf{Bi-GRU: Information Communication along Time.}
Lastly, LATTE employs a recurrent neural network to model the sequential dependency between visits in order to learn visit representation, upon which phenotype incidents are predicted at each visit.
To enable both prior and future visit information to be utilized for the prediction at the current visit, one Bi-GRUs layer is employed for the sequential modeling. Bi-GRUs receives the patient's longitudinal visit embedding vectors $\bgf(\bD_i^t)$ from \textbf{CR}, along with their corresponding attention values $\psi_t\circ \bgF(\Dscr_i)$ learned by \textbf{VAN}, and outputs the representation of each visit
    \begin{align*}
    \bgG\{\bgF(\Dscr_i), \bgps(\Dscr_i)\} = &
    \mathrm{BiGRU}_{\bgG}\{\bgf(\bD_i^1) \psi_1\circ \bgF(\Dscr_i),\dots,\bgf(\bD_i^{T_i}) \psi_{T_i}\circ \bgF(\Dscr_i)\} \\
    = & \{\bF(1,\Dscr_i),\dots,\bF(T_i,\Dscr_i)\} \\
    = & \{{\Upsilon_i(1),\dots,\Upsilon_i(T_i)}\},
    \end{align*}
where $\Upsilon_i(t)$ denotes the visit representation of patient $i$ at visit $t$.

\subsubsection{Loss Functions for Semi-supervised Learning}\label{DLloss}

For a binary outcome $Y_i^t$, the pooled cross entropy loss across patients and visits would be the typical choice for model training, as done in SAMGEP \cite{ahuja2021samgep} and RETAIN \cite{choi2016retain}. However, the simple loss function has two major limitations for incident phenotyping: (i)
the outcomes $Y_i^t$ only contribute to the prediction model at time $t$, ignoring the longitudinal nature of the outcome-feature pairs; and (ii) the fact that there is no guarantee on the monotonicity/smoothness of the prediction model
may compromise its interpretability. For example, the cumulative incidence rate may be expected to be non-decreasing over time.
Moreover, a large size of training data is typically required to effectively utilize the capacity of the complex deep-learning model. This poses a major concern as it is very costly to generate a large quantity of gold-standard outcomes.
To address these challenges,
we propose a two-step semi-supervised training strategy combining information from outcomes $Y_i^t$ and silver-standard labels $S_i^t$ through supervised and unsupervised kernel-weighted losses $\Lcal_{\SL}$ and $\Lcal_{\UL}$ with penalty terms $\Pcal_{\SL}$ and $\Pcal_{\UL}$, which encourages the model's monotonicity/smoothness.

Next, we introduce the loss function components $\Lcal_{\SL}$,  $\Lcal_{\UL}$ and $\Pcal$ used for the two-step label efficient training strategy described in Section \ref{sec_Label_efficient_learning}.

\noindent \textbf{Kernel-weighted losses: Incorporating Distance to Incidence.}
The outcomes around a given time $t$ may provide useful modality information for the prediction model at that time $t$. For cumulative incidence, the features $\bD_i^t$ may follow different patterns depending on the \emph{distance} between $t$ and the onset time.
We translate the distance factor into a kernel weighting, where
\begin{align*}
w_{Y,i}^t = w_{\min} + \exp\{-d_{Y,i}(t)^2/(2h^2)\}, \,
d_{Y,i}(t) = \min\{|u-t|: Y_i^u = 1\}, \\
w_{S,i}^t = w_{\min} + \exp\{-d_{S,i}(t)^2/(2h^2)\}, \,
d_{S,i}(t) = \min\{|u-t|: S_i^u \ge \kappa\}.
\end{align*}
Here, $h$ is a bandwidth hyper-parameter, $\kappa$ is a threshold for silver-standard label $S_i^u$ above which incidence $Y_i^u$ is likely active, and we set $\min\{\emptyset\} = +\infty$.
A minimum weight of $w_{\min}$ ensures the stability of the loss.
With standardized kernel weighting,
we construct the supervised and unsupervised loss functions,
\begin{align}
\Lcal_{\SL}(\bgTh) = & -\frac{1}{n}\sum_{i=1}^{n} \frac{\sum_{t=1}^{T_i} w_{Y,i}^t [Y^t_i \log\{G_Y\circ \bF(t,\Dscr_i)\}+(1-Y^t_i) \log\{1-G_Y\circ \bF(t,\Dscr_i)\}]}{\sum_{t'=1}^{T_i} w_{Y,i}^{t'}}  \notag \\
\Lcal_{\UL}(\bgTh) = & -\frac{1}{N}\sum_{i=1}^{N} \frac{\sum_{t=1}^{T_i} w_{S,i}^t [S^t_i \log\{G_S\circ \bF(t,\Dscr_i)\}+(1-S^t_i) \log\{1-G_S\circ \bF(t,\Dscr_i)\}]}{\sum_{t'=1}^{T_i} w_{S,i}^{t'}}.
\label{eq:loss_predict}
\end{align}

\noindent \textbf{Penalty: Regularization toward Monotonicity/smoothness.}
Depending on the type of incidence studied,
we construct two penalty terms.
We use the full cohort for the construction of penalties because 1) no outcome information is needed;
and 2) the monotonicity/smoothness is expected for prediction over the full cohort.
For the prediction of the cumulative incidence rate
with $Y_i^t$ a counting process with at most one jump, we impose a penalty that encourages the longitudinal prediction to be non-decreasing across time,
\begin{align}
% \Pcal_{\CUM,\SL}(\bgTh) = \frac{1}{n}\sum_{i=1}^n\frac{1}{T_i-1}\sum_{t=1}^{T_i-1}
% \max\{ G_Y\circ \bF(t,\Dscr_i) -  G_Y\circ \bF(t+1,\Dscr_i) ,0\}, \notag \\
\Pcal_{\CUM}(\bgTh) = \frac{1}{N}\sum_{i=1}^N\frac{1}{T_i-1}\sum_{t=1}^{T_i-1}
\max\{ G_S\circ \bF(t,\Dscr_i) -  G_S\circ \bF(t+1,\Dscr_i) ,0\}. \label{eq:loss_increase}
\end{align}
For the prediction of recurrent incidence rate, with $Y_i^t$ episodically shifting between $0$ and $1$, we impose a penalty that regulates the longitudinal prediction to insure its smoothness over time.

\begin{align}
% \Pcal_{\REC,\SL}(\bgTh) = \frac{1}{n}\sum_{i=1}^n\frac{1}{T_i-1}\sum_{t=1}^{T_i-1}
% \|\bF(t,\Dscr_i) - \bF(t+1,\Dscr_i)\|_2, \notag \\
\Pcal_{\REC}(\bgTh) = \frac{1}{N}\sum_{i=1}^N\frac{1}{T_i-1}\sum_{t=1}^{T_i-1}
\|\bF(t,\Dscr_i) - \bF(t+1,\Dscr_i)\|_2. \label{eq:loss_smooth}
\end{align}
We denote $\Pcal$ the penalty
chosen between $\Pcal_{\CUM}$ and $\Pcal_{\REC}$ depending on the type of incidence studied.

\subsubsection{Label-efficient Training of LATTE}
\label{sec_Label_efficient_learning}
LATTE is built upon deep neural networks, which in general heavily depend on large-scale annotations. In order to address the possible over-fitting under a small set of gold-standard labels, we rely on scalable silver-standard labels upon which LATTE is optimized in two steps: unsupervised pre-training and semi-supervised joint training.

\noindent \textbf{Unsupervised Pre-training.}
In the first step, LATTE is pre-trained using the silver labels. As shown in Figure \ref{figure:pipeline}, instead of sharing the same predictor $G_Y\circ \bF$ that predicts gold-standard labels, we attach an additional silver predictor $G_S\circ \bF$ upon the visit representation to predict the silver labels.
We pre-train all deep-learning model parameters $\bgTh$
excluding $G_Y$
using the unsupervised loss $\Lcal_{\UL}$ and the penalty $\Pcal$. We then denote the resulting loss
 \begin{align}
\Lcal_{\PT}(\bgTh)=\Lcal_{\UL}(\bgTh)+\lambda \Pcal(\bgTh), \\
\hat{\bgTh}_{\PT} = \argmin_{\bgTh} \Lcal_{\PT}(\bgTh),
\label{eq:silver_loss}
\end{align}
\noindent where $\lambda$ is a hyper-parameter determining the level of penalization. The concept re-weighting $\Kcal_{\CR}(\bge_j, \bge_{\varsigma})$ learned in the pre-training can then be used in another round of feature selection because it reflects the relevance of the features $\bD_i^t$ to the silver-standard labels $S_i^t$.

\noindent \textbf{Semi-supervised Joint Training.}
In the second step, LATTE performs semi-supervised fine-tuning of the model using both the gold-standard labels and the silver labels. The use of separate predictors for those two types of labels aims to prevent the potential poor quality of silver-standard labels from deteriorating the learning of gold-standard labels. The final training objective then becomes:
 \begin{gather}
\Lcal_{\SSL}(\triangle \bgTh)=\Lcal_{\SL}(\hat{\bgTh}_{\PT} + \triangle\bgTh)+ \gamma \Lcal_{\UL}(\hat{\bgTh}_{\PT} + \triangle\bgTh) + \lambda \Pcal(\hat{\bgTh}_{\PT} + \triangle\bgTh), \notag \\
\hat{\bgTh}_{\LATTE} = \hat{\bgTh}_{\PT} + \argmin_{\triangle \bgTh} \Lcal_{\SSL}(\triangle \bgTh),
\label{eq:final_loss}
\end{gather}
\noindent where $\gamma$ balances the contribution of gold-standard labels and silver-standard labels. \textbf{CR} model $\Kcal_{\CR}$, \textbf{VAN} modules $(\bQcal_{\VAN}, \bKcal_{\VAN})$ and the GRUs model $\bgG$ are jointly optimized according to the combination of the three objective functions.
The output transformations $G_Y$ and $G_S$
are optimized according to their involvements in $\Lcal_{\SL}(\bgTh)$ and $\Lcal_{\UL}(\bgTh)$ respectively.

\subsection{Enhancing Cross-site Portability}
\label{sec_portability}

In this section, we further strengthen LATTE's cross-site portability via contrastive representation learning.
We consider three aspects of data shift: a) concept utilization bias, b) visit frequency, and c) visit phase. Utilization bias is the fact that different institutions tend to have different preferences on concept utilization, especially on medications. Visit frequency shifts reflect the idea that different patients visit hospitals at different frequencies. Visit phase shift is caused by the fact that patients tend to visit and be discharged from hospitals during different stages of the phenotype, causing longitudinal EHR data to observe patients at various phenotype phases. For example, some patients may have already developed heart failure by the time of their first EHR visit while other patients may not have signs of the condition yet.

To enhance the robustness of LATTE toward these data shifts, we construct a robustness measure based on synthetic data that will be incorporated into the loss functions.
In order to mimic the above three data shifts, we generate the synthetic data
$$
\widetilde{\Lscr} = \{(\widetilde{Y}_{i,k}^t,\widetilde{\bD}_{i,k}^t): t=1,\dots,\widetilde{T}_{i,k}, \, i = 1,\dots,N, \,
k = 1,\dots, M
\}
$$ from labeled data $\Lscr$ using the bellow strategy.

\begin{enumerate}
    \item Placing a random Gaussian noise and a random corruption on concept counts to randomly set some concept counts to zero. We sample
    $\varepsilon_{i,k,j}^t \sim \mathcal{N}(1 , 0.05)$, $\delta_{i,k,j}^t \sim Bern(0.9)$ and add noise to the features,
$$
\widetilde{D}_{i,k,j}^{t}
= \delta_{i,k,j}^t(\overline{D}_{i,k}^{t}+\varepsilon_{i,k,j}^t).
$$
    \item Aggregating the visit sequence with varied time windows to mimic varied visit frequency. With an integer hyper-parameter $a_{\mathrm{mid}}$, we cycle $\widetilde{A}_{i,k}$ through $a_{\mathrm{mid}}-1,a_{\mathrm{mid}},a_{\mathrm{mid}}+1$.
    We aggregate the original $\check{T}_{i,k}$ visits into $\widetilde{T}_{i,k}=\lceil\check{T}_{i,k} / \widetilde{A}_{i,k}\rceil$ visits, where $\lceil x\rceil$ denotes the smallest integer bigger than $x$.
    A new visit at time $t$ thus combines the previous visits from
    $\epsilon_t = (t-1)\widetilde{A}_{i,k}+1$
    to $\nu_t = min\{t\widetilde{A}_{i,k},\check{T}_{i,k}\}$
    by taking the maximal outcome and summing over features
$$
\widetilde{Y}_{i,k}^t = \max_{u:\epsilon_t\le u \le \nu_t}\check{Y}_{i,k}^{u}, \,
\overline{\bD}_{i,k}^{t}
= \sum_{u = \epsilon_t}^{\nu_t} \check{\bD}_{i,k}^{u}, \,
t = 1,2,\dots,\widetilde{T}_{i,k}.
$$
    \item Randomly shifting the starting visit before the incidents of the EHR sequences to obtain sub-sequence samples.
    We randomly sample a truncation time $L_{i,k}$ from $\{0,1,\dots, min(T_i,L_{\max})\}$ with equal probability.  Here $L_{\max}$ is an integer hyper-parameter for maximal truncation time. We truncate the original data as
$$
\check{T}_{i,k} =  T_i - L_{i,k}, \,
(\check{Y}_{i,k}^{t},\check{\bD}_{i,k}^{t}) = (Y_{i}^{t-L_{i,k}},\bD_{i}^{t-L_{i,k}}), \,
t = 1,\dots,\check{T}_{i,k}.
$$

\end{enumerate}

% According to the design of synthetic data,
% the data at time $t$ for synthetic observation $(i,k)$
% is originated from original observation $i$ for times
% $$
% \Omega_{i,k}(t) = \{L_{i,k}+t\widetilde A_{i,k}+1, \dots, L_{i,k}+min\{(t+1)\widetilde{A}_{i,k},\widetilde{T}^{(1)}_{i,k}\}\}.
% $$
We construct a concordance of $\bF$ between $\Lscr$ and $\widetilde{\Lscr}$ to measure the robustness of the model towards data shifts,
$$
\Lcal_{\CC}(\bgTh)
= \frac{1}{N^2M}
\sum_{i'=1}^N
\sum_{i=1}^N\sum_{k=1}^M\sum_{t'=1}^{T_{i'}}
\sum_{t=1}^{\widetilde{T}_{i,k}}
\frac{
\max\left\{ (-1)^{\mathrm{I}(Y_{i'}^{t'} =\widetilde{Y}_{i,k}^t)}(c-\|\bF(t',\Dscr_{i'})-\bF(t,\widetilde{\Dscr}_{i,k})\|_2) ,0 \right\}}{T_{i'}\widetilde{T}_{i,k}},
$$
where $c$ is a tolerance for the contrast between positive and negative visits.  $\Lcal_{\CC}$ encourages the representation of positive visits to cluster together and keep at least a distance of $c$ from negative visits. Incorporating  $\Lcal_{\CC}$ with a hyper weighting parameter $\kappa$, the cross-site portable semi-supervised loss function becomes
$$
\Lcal_{\CP}( \bgTh)=\Lcal_{\SL}(\bgTh)+ \gamma \Lcal_{\UL}(\bgTh)  + \lambda \Pcal(\bgTh)
+ \kappa \Lcal_{\CC}(\bgTh).
$$

\section{Discussion}
This study proposes a computational framework based on semi-supervised learning for label-efficient incident phenotyping from longitudinal EHR data. Specifically, we develop and validate the proposed architecture named LATTE that identifies phenotype incident timings by learning to focus on the incident-indicative input features and patient visits and modeling the sequential dependency among visits. LATTE does not assume intensive inputs from experts for feature engineering and
can perform feature selection from large-scale EHR features in a data-driven manner.
It aims to be robust and scalable to all phenotypes with high clinical interpretation and directly portable across multiple clinical sites.
Experimental results on the three representative phenotypes show that LATTE identifies phenotype incident timings with high label efficiency. Particularly, LATTE consistently shows significant advantages over existing methods when the annotation size is small or the visit dependency is complicated as in the multiple sclerosis relapse.

While the identification of binary phenotypes using EHR data is pervasive in the literature, the
identification of longitudinal phenotype processes, or incident times, remains underdeveloped. Incident timings are expected to facilitate many EHR-related studies including supporting real-world evidence on the efficacy or safety of therapeutic drugs or intervention procedures. The incident timings determine baseline eligibility, define time-to-event outcomes and facilitate patient survival analysis. Further, the phenotype incident timings pave the way to mining temporal dependencies among phenotypes and uncovering disease progression trajectories.

There are several directions for future studies, especially in regard to novel phenotypes. First, improved concept embedding vectors can be used by combining the EHR embedding vectors with those from deep language models that are trained based on expert-curated knowledge. Such embedding vectors may not be available for novel diseases as there would be limited EHR data recording their clinical co-occurrences or relationship with the rest of healthcare concepts. Further, these EHR-derived embeddings tend to be dominated by common phenotypes and could be ill-performing for rare ones.  Secondly, instead of using one major surrogate for the unsupervised pre-training and semi-supervised co-training, we expect improved performances by mining the clinical relationship between target phenotypes and multiple weak surrogate concepts because, for a novel disease such as long COVID-19, the major concept that is predictive of it could be unknown or does not exist.

\section{Experimental Procedures}
\subsection{Resource availability}
\subsubsection{Lead contact}
Further information and requests for resources and reagents should be
directed to and will be fulfilled by the lead contact, Tianxi Cai (tcai@hsph.harvard.edu).

\subsubsection{Materials availability}
This study did not generate any physical materials.

\subsubsection{Data and code availability}
The clinical data reported in this study cannot be deposited in a public repository
due to regulations on protected health information (PHI). All original codes with simulated example data have been
deposited at https://github.com/celehs/LATTE/tree/peerreview. We also provide the pre-trained incident phenotyping models for the 3 representative phenotypes and the embedding vectors and weights of selected EHR concepts upon request. Any information required to reanalyze the data reported in this paper is available from the lead contact upon reasonable request.

\subsection{Implementation Details}
The concept reweighting module consists of three layers with the first layer containing two branches, each with 64 units, and the fusion layer with 32 units, followed by the output layer with one unit. The  visit attention network uses 64-dimensional query and key vectors, and the GRUs layer contains 128 units. For the silver-standard label construction defined in \eqref{eq:silver_label_cum} and \eqref{eq:silver_label_rec}, we set $\tau=0.1$ and $\alpha=0.2$. For cross-site portability, we set the distance tolerance $c=10$ and visit shifts $L_{\max}=7$. In the final training objective defined in \eqref{eq:final_loss}, we set $\gamma=0.1$, $\lambda=0.5$, and $\kappa=0.1$ to balance those training objectives. The whole model in optimized end-to-end using Adam \cite{kingma2014adam} with a learning rate of 0.001.

\section*{Author Contributions}
Conceptualization: T.C.; Methodology: J.W., T.C.; Data processing and analysis: J.W., J.H., B.C., V.M.C., V.A.P., W.D., Y.H., L.C.; Project administration: K.C.; Writing: J.W., T.C., J.H., B.C., K.P.L. Y.H.; Guarantors: T.C.; Approval of final manuscript: all authors.

\section*{Funding}
Part of this research is based on data from the Million Veteran Program, Office of Research and Development, Veterans Health Administration, and is supported by award \#MVP000.

\bibliographystyle{natbib}
\bibliography{reference}

\begin{thebibliography}{}

\bibitem[Ahuja {\em et~al.}(2020)Ahuja, Zhou, He, Sun, Castro, Gainer, Murphy,
  Hong, and Cai]{ahuja2020surelda}
Ahuja, Y.  {\em et~al.} (2020).
\newblock surelda: A multidisease automated phenotyping method for the
  electronic health record.
\newblock {\em Journal of the American Medical Informatics Association\/}, {\bf
  27}(8), 1235--1243.

\bibitem[Ahuja {\em et~al.}(2022)Ahuja, Wen, Hong, Xia, Huang, and
  Cai]{ahuja2021samgep}
Ahuja, Y.  {\em et~al.} (2022).
\newblock A semi-supervised adaptive markov gaussian embedding process (samgep)
  for prediction of phenotype event times using the electronic health record.
\newblock {\em Scientific reports\/}, {\bf 12}(1), 1--12.

\bibitem[Ananthakrishnan {\em et~al.}(2013)Ananthakrishnan, Cai, Savova, Cheng,
  Chen, Perez, Gainer, Murphy, Szolovits, Xia, {\em
  et~al.}]{ananthakrishnan2013improving}
Ananthakrishnan, A.~N.  {\em et~al.} (2013).
\newblock Improving case definition of crohn's disease and ulcerative colitis
  in electronic medical records using natural language processing: a novel
  informatics approach.
\newblock {\em Inflammatory bowel diseases\/}, {\bf 19}(7), 1411--1420.

\bibitem[Arnab {\em et~al.}(2021)Arnab, Dehghani, Heigold, Sun,
  Lu{\v{c}}i{\'c}, and Schmid]{arnab2021vivit}
Arnab, A.  {\em et~al.} (2021).
\newblock Vivit: A video vision transformer.
\newblock In {\em Proceedings of the IEEE/CVF International Conference on
  Computer Vision\/}, pages 6836--6846.

\bibitem[Badger {\em et~al.}(2019)Badger, LaRose, Mayer, Bashiri, Page, and
  Peissig]{badger2019machine}
Badger, J.  {\em et~al.} (2019).
\newblock Machine learning for phenotyping opioid overdose events.
\newblock {\em Journal of biomedical informatics\/}, {\bf 94}, 103185.

\bibitem[Beam {\em et~al.}(2019)Beam, Kompa, Schmaltz, Fried, Weber, Palmer,
  Shi, Cai, and Kohane]{beam2019clinical}
Beam, A.~L.  {\em et~al.} (2019).
\newblock Clinical concept embeddings learned from massive sources of
  multimodal medical data.
\newblock In {\em PACIFIC SYMPOSIUM ON BIOCOMPUTING 2020\/}, pages 295--306.
  World Scientific.

\bibitem[Choi {\em et~al.}(2016)Choi, Bahadori, Sun, Kulas, Schuetz, and
  Stewart]{choi2016retain}
Choi, E.  {\em et~al.} (2016).
\newblock Retain: An interpretable predictive model for healthcare using
  reverse time attention mechanism.
\newblock {\em Advances in neural information processing systems\/}, {\bf 29}.

\bibitem[Chubak {\em et~al.}(2012)Chubak, Yu, Pocobelli, Lamerato, Webster,
  Prout, Ulcickas~Yood, Barlow, and Buist]{chubak2012administrative}
Chubak, J.  {\em et~al.} (2012).
\newblock Administrative data algorithms to identify second breast cancer
  events following early-stage invasive breast cancer.
\newblock {\em Journal of the National Cancer Institute\/}, {\bf 104}(12),
  931--940.

\bibitem[Devlin {\em et~al.}(2018)Devlin, Chang, Lee, and
  Toutanova]{devlin2018bert}
Devlin, J.  {\em et~al.} (2018).
\newblock Bert: Pre-training of deep bidirectional transformers for language
  understanding.
\newblock {\em arXiv preprint arXiv:1810.04805\/}.

\bibitem[Gamerman {\em et~al.}(2019)Gamerman, Cai, and
  Els{\"a}{\ss}er]{gamerman2019pragmatic}
Gamerman, V.  {\em et~al.} (2019).
\newblock Pragmatic randomized clinical trials: best practices and statistical
  guidance.
\newblock {\em Health Services and Outcomes Research Methodology\/}, {\bf
  19}(1), 23--35.

\bibitem[Hassett {\em et~al.}(2017)Hassett, Uno, Cronin, Carroll, Hornbrook,
  and Ritzwoller]{hassett2017detecting}
Hassett, M.~J.  {\em et~al.} (2017).
\newblock Detecting lung and colorectal cancer recurrence using structured
  clinical/administrative data to enable outcomes research and population
  health management.
\newblock {\em Medical care\/}, {\bf 55}(12), e88.

\bibitem[Hernandez-Boussard {\em et~al.}(2019)Hernandez-Boussard, Monda,
  Crespo, and Riskin]{hernandez2019real}
Hernandez-Boussard, T.  {\em et~al.} (2019).
\newblock Real world evidence in cardiovascular medicine: ensuring data
  validity in electronic health record-based studies.
\newblock {\em Journal of the American Medical Informatics Association\/}, {\bf
  26}(11), 1189--1194.

\bibitem[Hong {\em et~al.}(2021)Hong, Rush, Liu, Zhou, Sun, Sonabend, Castro,
  Schubert, Panickan, Cai, {\em et~al.}]{hong2021clinical}
Hong, C.  {\em et~al.} (2021).
\newblock Clinical knowledge extraction via sparse embedding regression (keser)
  with multi-center large scale electronic health record data.
\newblock {\em NPJ digital medicine\/}, {\bf 4}(1), 1--11.

\bibitem[Hou {\em et~al.}(????)Hou, Chan, Wang, and Cai]{HouChanWangCaiBx21}
Hou, J.  {\em et~al.} (????).
\newblock Risk prediction with imperfect survival outcome information from
  electronic health records.
\newblock {\em Biometrics\/}, {\bf n/a}(n/a).

\bibitem[Hou {\em et~al.}(2021)Hou, Kim, Cai, Dahal, Weiner, Chitnis, Cai, and
  Xia]{hou2021comparison}
Hou, J.  {\em et~al.} (2021).
\newblock Comparison of dimethyl fumarate vs fingolimod and rituximab vs
  natalizumab for treatment of multiple sclerosis.
\newblock {\em JAMA network open\/}, {\bf 4}(11), e2134627--e2134627.

\bibitem[Hou {\em et~al.}(2022)Hou, Zhao, Cai, Beaulieu-Jones, Seyok, Dahal,
  Yuan, Xiong, Bonzel, Fox, {\em et~al.}]{hou2022temporal}
Hou, J.  {\em et~al.} (2022).
\newblock Temporal trends in clinical evidence of 5-year survival within
  electronic health records among patients with early-stage colon cancer
  managed with laparoscopy-assisted colectomy vs open colectomy.
\newblock {\em JAMA network open\/}, {\bf 5}(6), e2218371--e2218371.

\bibitem[Huang {\em et~al.}(2021)Huang, Cai, Weber, He, Dahal, Hong, Hou,
  Seyok, Cagan, DiCarli, {\em et~al.}]{huang2021association}
Huang, S.  {\em et~al.} (2021).
\newblock The association between inflammation, incident heart failure, and
  heart failure subtypes in patients with rheumatoid arthritis.
\newblock {\em Arthritis Care \& Research\/}.

\bibitem[Kingma and Ba(2014)Kingma and Ba]{kingma2014adam}
Kingma, D.~P. and Ba, J. (2014).
\newblock Adam: A method for stochastic optimization.
\newblock {\em arXiv preprint arXiv:1412.6980\/}.

\bibitem[Kirby {\em et~al.}(2016)Kirby, Speltz, Rasmussen, Basford, Gottesman,
  Peissig, Pacheco, Tromp, Pathak, Carrell, {\em et~al.}]{kirby2016phekb}
Kirby, J.~C.  {\em et~al.} (2016).
\newblock Phekb: a catalog and workflow for creating electronic phenotype
  algorithms for transportability.
\newblock {\em Journal of the American Medical Informatics Association\/}, {\bf
  23}(6), 1046--1052.

\bibitem[Kohane {\em et~al.}(2012)Kohane, Churchill, and
  Murphy]{kohane2012translational}
Kohane, I.~S.  {\em et~al.} (2012).
\newblock A translational engine at the national scale: informatics for
  integrating biology and the bedside.
\newblock {\em Journal of the American Medical Informatics Association\/}, {\bf
  19}(2), 181--185.

\bibitem[Lee {\em et~al.}(2020)Lee, Rashbass, and Van~der
  Schaar]{lee2020outcome}
Lee, C.  {\em et~al.} (2020).
\newblock Outcome-oriented deep temporal phenotyping of disease progression.
\newblock {\em IEEE Transactions on Biomedical Engineering\/}, {\bf 68}(8),
  2423--2434.

\bibitem[Levy and Goldberg(2014)Levy and Goldberg]{levy2014neural}
Levy, O. and Goldberg, Y. (2014).
\newblock Neural word embedding as implicit matrix factorization.
\newblock {\em Advances in neural information processing systems\/}, {\bf 27}.

\bibitem[Liao {\em et~al.}(2010)Liao, Cai, Gainer, Goryachev, Zeng-treitler,
  Raychaudhuri, Szolovits, Churchill, Murphy, Kohane, {\em
  et~al.}]{liao2010electronic}
Liao, K.~P.  {\em et~al.} (2010).
\newblock Electronic medical records for discovery research in rheumatoid
  arthritis.
\newblock {\em Arthritis care \& research\/}, {\bf 62}(8), 1120--1127.

\bibitem[Liao {\em et~al.}(2019)Liao, Sun, Cai, Link, Hong, Huang, Huffman,
  Gronsbell, Zhang, Ho, {\em et~al.}]{liao2019high}
Liao, K.~P.  {\em et~al.} (2019).
\newblock High-throughput multimodal automated phenotyping (map) with
  application to phewas.
\newblock {\em Journal of the American Medical Informatics Association\/}, {\bf
  26}(11), 1255--1262.

\bibitem[Liu {\em et~al.}(2015)Liu, Wang, Hu, and Xiong]{liu2015temporal}
Liu, C.  {\em et~al.} (2015).
\newblock Temporal phenotyping from longitudinal electronic health records: A
  graph based framework.
\newblock In {\em Proceedings of the 21th ACM SIGKDD international conference
  on knowledge discovery and data mining\/}, pages 705--714.

\bibitem[Mikolov {\em et~al.}(2013)Mikolov, Sutskever, Chen, Corrado, and
  Dean]{mikolov2013distributed}
Mikolov, T.  {\em et~al.} (2013).
\newblock Distributed representations of words and phrases and their
  compositionality.
\newblock {\em Advances in neural information processing systems\/}, {\bf 26}.

\bibitem[Miotto {\em et~al.}(2016)Miotto, Li, Kidd, and Dudley]{miotto2016deep}
Miotto, R.  {\em et~al.} (2016).
\newblock Deep patient: an unsupervised representation to predict the future of
  patients from the electronic health records.
\newblock {\em Scientific reports\/}, {\bf 6}(1), 1--10.

\bibitem[Murphy {\em et~al.}(2006)Murphy, Mendis, Berkowitz, Kohane, and
  Chueh]{murphy2006integration}
Murphy, S.~N.  {\em et~al.} (2006).
\newblock Integration of clinical and genetic data in the i2b2 architecture.
\newblock In {\em AMIA Annual Symposium Proceedings\/}, volume 2006, page 1040.
  American Medical Informatics Association.

\bibitem[Newton {\em et~al.}(2013)Newton, Peissig, Kho, Bielinski, Berg,
  Choudhary, Basford, Chute, Kullo, Li, {\em et~al.}]{newton2013validation}
Newton, K.~M.  {\em et~al.} (2013).
\newblock Validation of electronic medical record-based phenotyping algorithms:
  results and lessons learned from the emerge network.
\newblock {\em Journal of the American Medical Informatics Association\/}, {\bf
  20}(e1), e147--e154.

\bibitem[Nicola {\em et~al.}(2005)Nicola, Maradit-Kremers, Roger, Jacobsen,
  Crowson, Ballman, and Gabriel]{nicola2005risk}
Nicola, P.~J.  {\em et~al.} (2005).
\newblock The risk of congestive heart failure in rheumatoid arthritis: a
  population-based study over 46 years.
\newblock {\em Arthritis \& Rheumatism: Official Journal of the American
  College of Rheumatology\/}, {\bf 52}(2), 412--420.

\bibitem[Roden {\em et~al.}(2008)Roden, Pulley, Basford, Bernard, Clayton,
  Balser, and Masys]{roden2008development}
Roden, D.~M.  {\em et~al.} (2008).
\newblock Development of a large-scale de-identified dna biobank to enable
  personalized medicine.
\newblock {\em Clinical Pharmacology \& Therapeutics\/}, {\bf 84}(3), 362--369.

\bibitem[Uno {\em et~al.}(2018)Uno, Ritzwoller, Cronin, Carroll, Hornbrook, and
  Hassett]{uno2018determining}
Uno, H.  {\em et~al.} (2018).
\newblock Determining the time of cancer recurrence using claims or electronic
  medical record data.
\newblock {\em JCO clinical cancer informatics\/}, {\bf 2}, 1--10.

\bibitem[Vaswani {\em et~al.}(2017)Vaswani, Shazeer, Parmar, Uszkoreit, Jones,
  Gomez, Kaiser, and Polosukhin]{vaswani2017attention}
Vaswani, A.  {\em et~al.} (2017).
\newblock Attention is all you need.
\newblock {\em Advances in neural information processing systems\/}, {\bf 30}.

\bibitem[Yang {\em et~al.}(2022)Yang, Varghese, Stephenson, Tu, and
  Gronsbell]{ocac216}
Yang, S.  {\em et~al.} (2022).
\newblock {Machine learning approaches for electronic health records
  phenotyping: a methodical review}.
\newblock {\em Journal of the American Medical Informatics Association\/}.
\newblock ocac216.

\bibitem[Yu {\em et~al.}(2013)Yu, Cai, and Cai]{yu2013nile}
Yu, S.  {\em et~al.} (2013).
\newblock Nile: fast natural language processing for electronic health records.
\newblock {\em arXiv preprint arXiv:1311.6063\/}.

\bibitem[Yu {\em et~al.}(2018)Yu, Ma, Gronsbell, Cai, Ananthakrishnan, Gainer,
  Churchill, Szolovits, Murphy, Kohane, {\em et~al.}]{yu2018enabling}
Yu, S.  {\em et~al.} (2018).
\newblock Enabling phenotypic big data with phenorm.
\newblock {\em Journal of the American Medical Informatics Association\/}, {\bf
  25}(1), 54--60.

\bibitem[Zhou {\em et~al.}(2022)Zhou, Gan, Shi, Patwari, Rush, Bonzel,
  Panickan, Hong, Ho, Cai, {\em et~al.}]{zhou2022multiview}
Zhou, D.  {\em et~al.} (2022).
\newblock Multiview incomplete knowledge graph integration with application to
  cross-institutional ehr data harmonization.
\newblock {\em Journal of Biomedical Informatics\/}, {\bf 133}, 104147.

\end{thebibliography}

\end{document}